\documentclass[journal]{IEEEtran}
\usepackage{cite}
\usepackage{graphicx}
\usepackage{amsmath}
\usepackage{amssymb}
\usepackage{subfigure}
\usepackage{stfloats}
\usepackage{algorithm}
\usepackage{algorithmicx}
\usepackage{algpseudocode}
\usepackage{url}
\usepackage{tabularx}
\usepackage{array}
\usepackage{multirow}
\usepackage{color}
\usepackage{booktabs}
\usepackage{enumerate}
\usepackage{breqn}
\usepackage{indentfirst}
\usepackage{bm}
\usepackage{mathtools}
\usepackage{xcolor}
\usepackage{svg}
\usepackage[export]{adjustbox}
\usepackage[linkcolor=black,citecolor=black,urlcolor=black,colorlinks=true]{hyperref}

\graphicspath{{fig/}}

\begin{document}
	\title{G\ensuremath{^2}VD Planner: Efficient Motion Planning With Grid-based Generalized Voronoi Diagrams}
	
	\author{\IEEEauthorblockN{Jian Wen, Xuebo Zhang,~\IEEEmembership{Senior Member,~IEEE,} Qingchen Bi, Hui Liu, Jing Yuan,~\IEEEmembership{Member,~IEEE,}\\ and Yongchun Fang,~\IEEEmembership{Senior Member,~IEEE}}
	\thanks{This work was supported in part by the National Natural Science Foundation of China under Grant 62293513/62293510, in part by Natural Science Foundation of Tianjin under Grant 22JCZDJC00810, and in part by the Fundamental Research Funds for the Central Universities. \emph{(Corresponding author: Xuebo Zhang.)}}
	\thanks{The authors are with the Institute of Robotics and Automatic Information System, College of Artificial Intelligence, Nankai University, Tianjin 300350, China, and also with the Tianjin Key Laboratory of Intelligent Robotics, Nankai University, Tianjin 300350, China (e-mail: zhangxuebo@nankai.edu.cn).}
	}

\maketitle

\begin{abstract}
	In this paper, an efficient motion planning approach with grid-based generalized Voronoi diagrams (G\ensuremath{^2}VD) is newly proposed for mobile robots. Different from existing approaches, the novelty of this work is twofold: 1) a new state lattice-based path searching approach is proposed, in which the search space is reduced to a novel Voronoi corridor to further improve the search efficiency; 2) an efficient quadratic programming-based path smoothing approach is presented, wherein the clearance to obstacles is considered to improve the path clearance of hard-constrained path smoothing approaches. We validate the efficiency and smoothness of our approach in various challenging simulation scenarios and outdoor environments. It is shown that the computational efficiency is improved by 17.1\% in the path searching stage, and path smoothing with the proposed approach is 6.6 times faster than an advanced sparse-banded structure-based path smoothing approach and 53.3 times faster than the popular timed-elastic-band planner. A video showing outdoor navigation on our campus is available at \url{https://youtu.be/iMXGthgvp58}.
\end{abstract}

\def\abstractname{Note to Practitioners}
\begin{abstract}
	This paper is motivated by the challenges of motion planning problems of mobile robots. An efficient motion planning approach called G\ensuremath{^2}VD planner is proposed by combining path searching, path smoothing, and time-optimal velocity planning. Extensive simulation and experimental results show the effectiveness of the proposed motion planning approach. However, the prediction information of dynamic obstacles is not incorporated in the proposed motion planner, thus the motion planner may be a bit sluggish in response to dynamic obstacles. Furthermore, we plan to integrate the intention/trajectory prediction of pedestrians/vehicles into the proposed framework to enhance the foreseeability of the motion planner.
\end{abstract}


\IEEEpeerreviewmaketitle

\section{Introduction}
\label{introduction}
\IEEEPARstart{M}{obile} robots have been widely applied in various fields including but not limited to intelligent inspection, logistics, and search-and-rescue applications \cite{chen2021integrated,bai2019efficient,bai2022group}. Autonomous navigation is the key technology for mobile robots to achieve full autonomy in these scenarios \cite{li2022sensing,song2023multi}, which typically consists of localization \cite{gao2023negl,wen2022tm3loc}, mapping \cite{wen2021bridging,wen2022roadside}, motion planning, and control, wherein motion planning plays an essential role in generating safe, smooth, and efficient motions \cite{latombe1991robot,choset2005principles}. Although plenty of works on motion planning of mobile robots have been put forward \cite{wang2022efficient,wang2022deep,zhou2021trajectory}, it is still challenging to design a motion planning approach that can ensure both efficiency, safety, and smoothness in complex environments \cite{bi2023cure,chi2018risk,wang2020neural,wang2020eb,zhang2023dual}.

\subsection{Path Searching}
Many path searching approaches have been proposed in terms of different theories \cite{lavalle2006planning}, which can be classified into three categories. The sampling-based planning algorithms such as the famous probabilistic roadmap (PRM) \cite{kavraki1996probabilistic} and rapidly-exploring random trees (RRTs) \cite{lavalle2001randomized} have gained popularity for the capability of efficient searching in the configuration space (C-space). However, they are limited by completeness and optimality, and even some excellent variants such as RRT* \cite{karaman2011sampling} can only guarantee asymptotic optimality. The classical artificial potential field (APF) algorithm \cite{khatib1986real} finds a feasible path by following the steepest descent of a potential field. However, the APF algorithm often suffers from the local minimum problem. In this paper, we focus on grid-based planning algorithms. Grid-based planning overlays a hyper-grid on the C-space and assumes each configuration is identified with the grid-cell center \cite{bai2019distributed}. Then, search algorithms are used to find a path from the start to the goal. Grid-based planning can always find a resolution-optimal path if it exists in the discrete search space, namely, completeness and optimality in the sense of resolution are guaranteed. However, these approaches depend on space discretization and do not perform fast as the environment dimension increases.

\subsection{Path Smoothing}
The path obtained by path searching approaches usually fails to meet the smoothness requirement for robot navigation and needs further smoothing \cite{ravankar2018path}. Two types of path smoothing approaches are investigated in this paper, namely, soft-constrained approaches and hard-constrained approaches.

\subsubsection{Soft-constrained Approaches} 
These approaches formulate path smoothing as a non-linear unconstrained optimization problem, wherein the constraints are considered in the optimization objective in the form of penalty cost terms \cite{rosmann2017kinodynamic,deray2019timed,smith2020egoteb}. Generally, the smoothness of the path and the clearance to obstacles are both taken into account. Then, gradient-based optimization algorithms are employed to solve the problem. Soft-constrained approaches utilize gradient information to push the path far from obstacles and can obtain a smooth path with a reasonable distance from obstacles. However, it is difficult for these approaches to guarantee the optimized path strictly satisfies the constraints. In addition, the optimization objective of soft-constrained approaches contains convex and non-convex terms, making these approaches suffer from local optima and time-consuming issues \cite{rosmann2017integrated}.

\subsubsection{Hard-constrained Approaches}
These approaches formulate path smoothing as a non-linear constrained optimization problem and can obtain a path that theoretically satisfies the constraints \cite{zhang2018autonomous,zhang2020optimization}. The optimization objective usually considers the smoothness or the length of the path and thus is convex. This convexity allows the problem to be solved efficiently in general \cite{zhu2015convex,liu2017convex}. However, hard-constrained approaches treat all free space equally, namely, distance from feasible paths to obstacles is ignored. As a result, the optimized path is often close to the obstacle and has poor safety.

\subsection{Contributions}
Motivated by the aforementioned limitations of existing works, an efficient motion planning approach called G\ensuremath{^2}VD planner is newly proposed. Specifically, a G\ensuremath{^2}VD is utilized to aid path searching and path smoothing to achieve better performance in terms of efficiency, safety, smoothness, etc.

\subsubsection{Path Searching}
Given the start and the goal, we first employ the A* search to find the shortest grid path in a G\ensuremath{^2}VD. This shortest Voronoi path contains the topological information of the search direction and provides rough guidance for the subsequent fine search. Along the Voronoi grid path, a region called \emph{Voronoi corridor} is proposed and constructed by adding a bounding box to each path pixel. Furthermore, the cost function of path searching is redesigned based on a potential called \emph{Voronoi field} to make the searched path keep a reasonable distance from obstacles. Finally, the A* search combined with motion primitives is utilized to finely search a kinematically feasible path within the Voronoi corridor. Different from the original state lattice-based path planner \cite{likhachev2009planning} that takes the whole grid map as the search space, the proposed approach reduces the search space to the Voronoi corridor, thus considerable time is saved for path searching. In addition, the certain clearance to obstacles provides sufficient optimization margin for further path smoothing.

\subsubsection{Path Smoothing}
Taking the path searched above as the reference path, an efficient quadratic programming (QP)-based path smoothing approach is proposed, wherein the smoothness of the path and the deviation from the reference path are both considered in the optimization objective. Our goal is to obtain a smooth path while minimizing the deviation between the optimized path and the reference path. Because the reference path searched above has a certain distance to obstacles, the clearance to obstacles is implicitly considered in the proposed approach, and the path clearance issue of such a hard-constrained approach is addressed. This is also the second reason why we introduce the Voronoi field and redesign the cost function of path searching to improve the path clearance of the searched path, in addition to providing wider optimization margin for path smoothing.

To summarize, the main contributions of this work are as follows:
\begin{enumerate}[\hspace{1em}1)]
	\item A new state lattice-based path searching approach is proposed, in which a novel Voronoi corridor is introduced to reduce the search space to significantly improve the search efficiency, along with a Voronoi potential constructed to make the searched path keep a reasonable distance from obstacles to provide sufficient optimization margin for further path smoothing.
	\item A new QP-based path smoothing approach is presented to efficiently smooth the searched path, wherein the clearance to obstacles is considered in the form of the penalty of the deviation from the safe reference path to address the path clearance issue of existing hard-constrained path smoothing approaches.
	\item Autonomous navigation is realized in outdoor environments. Extensive simulation and experimental evaluations are presented to validate the effectiveness of the proposed approach.
\end{enumerate}

The rest of this paper is organized as follows. We first review the related work in Section \ref{relatedwork}. The proposed G\ensuremath{^2}VD planner is detailed in Section \ref{approach}. Section \ref{implementationdetails} provides some implementation details. The results of simulations and experiments are presented in Sections \ref{simulations} and \ref{experiments}, respectively. Finally, this paper is concluded in Section \ref{conclusion}.

\section{Related work}
\label{relatedwork}
Grid-based planning obtains the resolution-optimal path by discretizing the C-space first and then using graph search algorithms to find the path. In \cite{bai2019distributed}, a novel grid modeling-based path planning approach is introduced, which has been successfully applied in the distributed multi-vehicle task assignment. However, original grid-based planning approaches only visit the centers of grid cells and produce piecewise-linear paths that do not generally satisfy the kinodynamic constraints of the robot. To address this problem, Pivtoraiko \emph{et al.} \cite{pivtoraiko2009differentially} propose the state lattice approach for graph construction. In particular, the connectivity between two nodes in the graph is built from a pre-designed motion primitive that fulfills the kinodynamic constraints of the robot. Based on the work \cite{pivtoraiko2009differentially}, Likhachev \emph{et al.} present a state lattice-based path planner by combining AD* search with motion primitives \cite{likhachev2009planning}, which has been successfully applied in the DARPA Urban Challenge \cite{ferguson2008motion}. In this paper, a new state lattice-based path planner is proposed, wherein the search space is reduced to a Voronoi corridor derived from a generalized Voronoi diagram (GVD) to further improve the search efficiency.

The use of GVDs has long been proposed in the context of robot motion planning. In \cite{ayawli2019mobile}, Ayawli \emph{et al.} propose a path planning approach for mobile robots using Voronoi diagrams and computational geometry technologies. In \cite{choset2000sensor}, Choset and Burdick use the GVD to derive skeletons of the free space and then search on the graph. However, the shortest Voronoi path may be far from the actual optimal path. In \cite{ziegler2008navigating}, Ziegler \emph{et al.} utilize the GVD to inform the A* search by constructing a heuristic where the cost of a search state is the sum of the straight-line path to the closest Voronoi edge and the shortest path along the GVD edges. However, the resulting heuristic cannot be guaranteed to be admissible since the cost of the shortest Voronoi path may be greater than that of the actual shortest path. In \cite{dolgov2010path}, Dolgov \emph{et al.} design a potential field based on the GVD for path smoothing. This potential allows precise navigation in narrow passages while also effectively pushing the robot away from obstacles in wider areas. Inspired by the Voronoi field, we redesign the cost function of the state lattice-based path planner to obtain a path with a reasonable distance from obstacles.

The path obtained by path searching approaches usually needs further smoothing. In \cite{cai2022human,suzuki2010smooth}, comparisons of path smoothing approaches in dynamic environments are presented. In \cite{dolgov2010path}, Dolgov \emph{et al.} present a conjugate gradient-based path smoothing approach to smooth the path generated by the hybrid A* algorithm. In \cite{wen2020effmop}, Wen \emph{et al.} propose a gradient-based local path smoothing approach for mobile robots, wherein the sparse-banded system structure of the underlying optimization problem is fully exploited to efficiently solve the problem. The above soft-constrained approaches utilize gradient information to push the path far from obstacles and can obtain a path with better safety. However, it is difficult for these approaches to guarantee the optimized path strictly satisfies the constraints. Recently, Zhou \emph{et al.} propose a dual-loop iterative anchoring path smoothing approach for autonomous driving \cite{zhou2020autonomous}, in which the nonlinear curvature constraint is linearized and sequential convex programming is used to efficiently solve the path optimization problem. Such a hard-constrained approach can theoretically guarantee that the obtained path strictly satisfies the constraints. However, distance from feasible paths to obstacles is ignored, which often results in the optimized path being close to obstacles and poor safety. In this paper, a new QP-based path smoothing approach is proposed, in which the clearance to obstacles is implicitly considered in the form of the penalty of the deviation from the safe reference path to improve the path clearance of hard-constrained approaches.

\begin{figure}[t]
	\centering
	\includegraphics[scale=0.29]{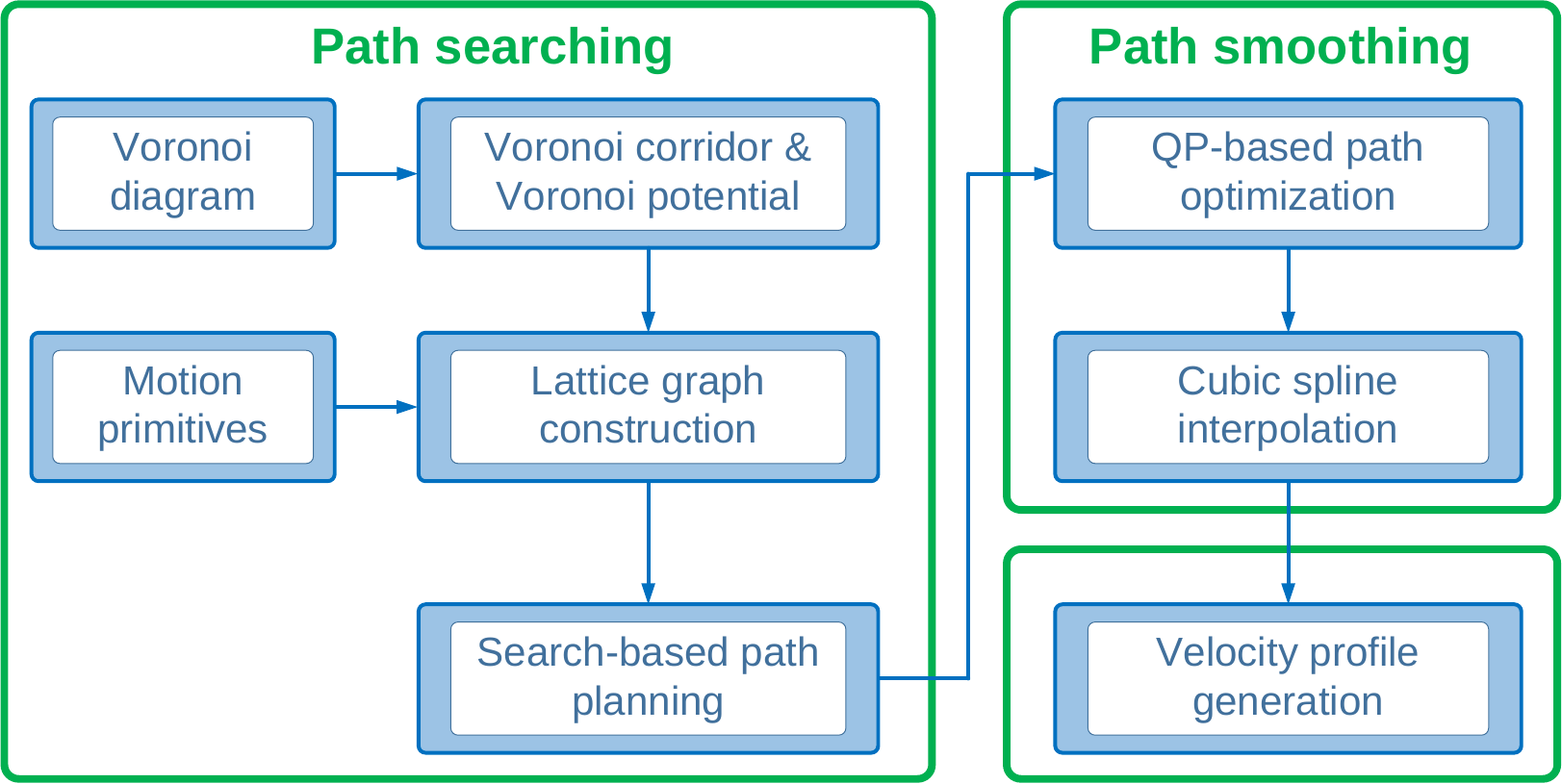}
	\caption{Flow chart of the proposed three-layer motion planning framework.}
	\label{fig:frameork}
\end{figure}

\section{Problem Definition}
\label{statement}
Let $ \mathcal{X} \subset \mathbb{R}^n $ be the state space. The obstacle space and the free space are denoted as $ \mathcal{X}_{\text{obs}} $ and $ \mathcal{X}_{\text{free}}=\mathcal{X} \backslash \mathcal{X}_{\text{obs}} $, respectively. The motion planning aims to find a feasible trajectory $\phi(t)$ such that
\begin{equation}
	\phi(t):[0, T] \rightarrow \mathcal{X}_{\text{free}}, \phi(0)=\boldsymbol{S}_{\text{start}}, \phi(T) =\boldsymbol{S}_{\text{goal}},
\end{equation}
where $t$ is the timestamp, $\boldsymbol{S}_{\text{start}}$ and $\boldsymbol{S}_{\text{goal}}$ denote the start and goal states, respectively. In this work, the state $\boldsymbol{S}$ is defined as a tuple of $\left(x,y,\theta\right)$, where $ \left(x, y\right) $ denotes the position of the robot in the world and $ \theta $ represents the heading of the robot.

Directly solving the mapping of $t \rightarrow \left(x,y,\theta\right)$  has a huge computational burden.  In consideration of the computational efficiency, the motion planning problem is typically decoupled as path planning and velocity planning \cite{kant1986toward}. In particular, the path planning solves the mapping of $s \rightarrow \left(x,y,\theta\right)$ and the velocity planning solves the mapping of $t \rightarrow s$, where $s$ is the accumulated distance along a given path. In this work, a new state lattice-based path searching approach combined with an efficient QP-based path searching approach is proposed to solve the mapping of $s \rightarrow \left(x,y,\theta\right)$, while the mapping of $t \rightarrow s$ is solved by our previously proposed time-optimal velocity planning algorithm \cite{zhang2018multilevel}.

\section{G\ensuremath{^2}VD Planner}
\label{approach}

In this paper, an efficient three-layer motion planning framework called G\ensuremath{^2}VD planner is carefully designed, which consists of path searching, path smoothing, and velocity planning, as shown in Fig. \ref{fig:frameork}. The path searching module is utilized to provide a safe reference path for the robot, and the path smoothing module combined with our previously proposed time-optimal velocity planning \cite{zhang2018multilevel} is employed to generate safe, smooth, and efficient motion commands. In this section, we will detail the newly proposed path searching and path smoothing modules.

\begin{figure}[t]
	\centering
	\includegraphics[scale=0.3]{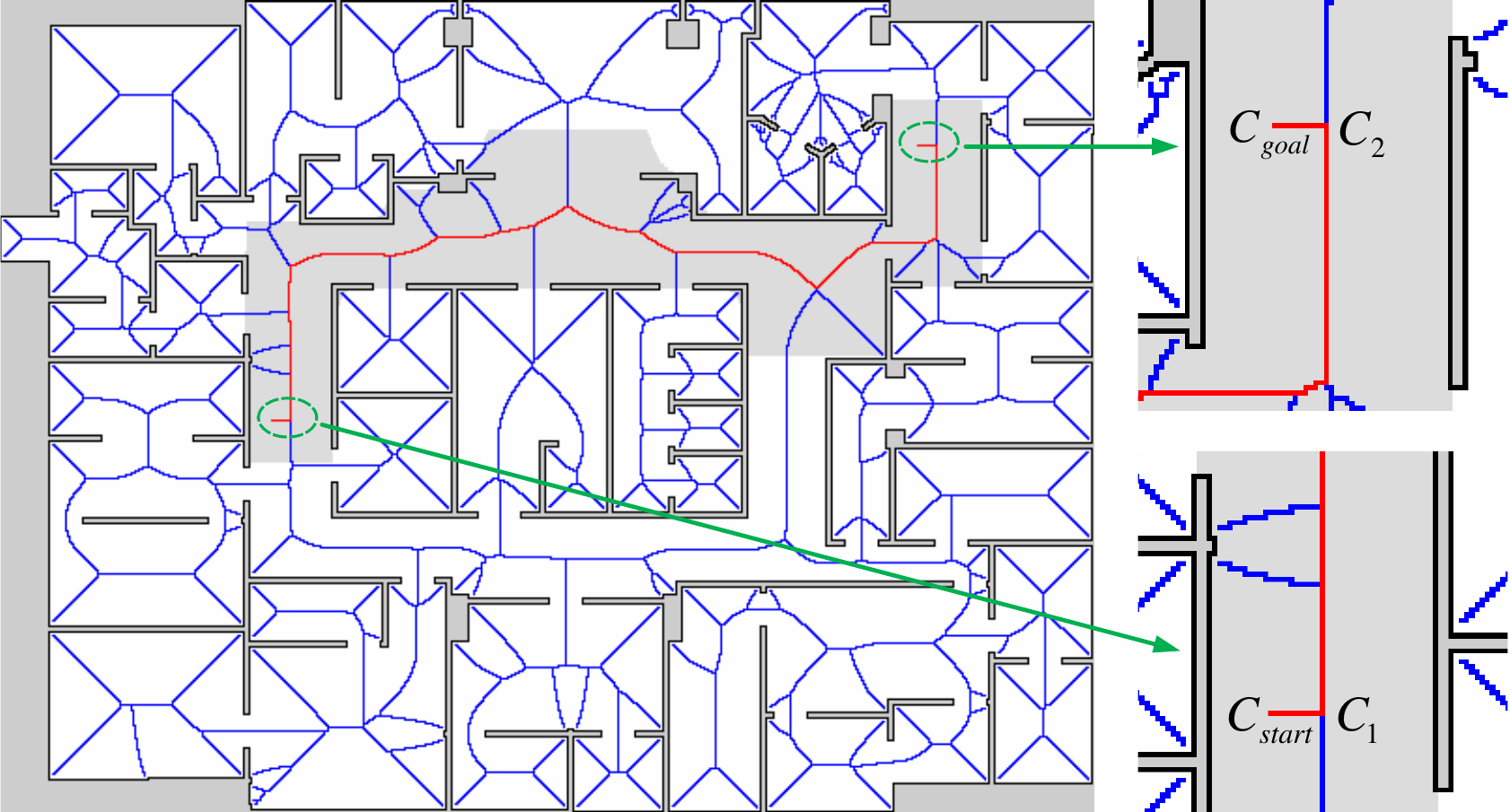}
	\caption{G\ensuremath{^2}VD of an office-like environment. The Voronoi edge pixels are colored in blue and red, and the shortest Voronoi path is colored in red. The gray region surrounding the Voronoi path denotes the Voronoi corridor.}
	\label{fig:voronoi}
\end{figure}

\subsection{Grid-based Generalized Voronoi Diagrams}
\label{update_mechanism}
The GVD is defined as the set of points in the free space to which the two closest obstacles have the same distance \cite{choset2000sensor}. In this paper, an efficient, incrementally updatable GVD construction algorithm presented in \cite{lau2013efficient} is utilized to convert the occupancy grid map to a GVD in discrete form. The core of this algorithm \cite{lau2013efficient} is to employ a dynamic variant of the brushfire algorithm \cite{kalra2009incremental} to incrementally compute Euclidean distances. Specifically, the G\ensuremath{^2}VD is firstly initialized based on a prior grid map, which is usually generated by simultaneous localization and mapping (SLAM) \cite{grisetti2007improved}. On top of the initial G\ensuremath{^2}VD, both moving dynamic obstacles and unknown static obstacles are detected by external sensors mounted on the robot. Movement, insertion, and deletion of obstacles change the states of individual cells in the G\ensuremath{^2}VD from free to occupied or vice versa. Newly occupied cells initiate ``lower'' wavefronts that update the distance to the closest obstacle of affected cells. Similarly, ``raise'' wavefronts start at newly freed cells and clear the distance entries of all cells whose closest obstacle is the deleted one. The processing of raised and lower wavefronts is interwoven and controlled by a single priority queue that sorts the enqueued cells by distance. Both the raised and lower wavefronts enqueue the neighbors of a processed cell to propagate the wavefronts. Readers can refer to \cite{lau2013efficient} for more details about this algorithm. Fig. \ref{fig:voronoi} illustrates the G\ensuremath{^2}VD of an office-like environment, in which the Voronoi edge pixels are colored in blue and red.

As mentioned above, the employed incrementally updatable G\ensuremath{^2}VD construction algorithm can deal with both moving dynamic obstacles and unknown static obstacles. Therefore, even if the prior grid map is not available, the robot can still observe the surrounding environment through external sensors and update the G\ensuremath{^2}VD in real-time. That is to say, the proposed framework can be theoretically applied in completely unknown scenarios. In practice, if a prior grid map containing the basic environmental structural information is available, it will enable the G\ensuremath{^2}VD planner to provide better global guidance.

\subsection{Voronoi Corridor}
After obtaining the G\ensuremath{^2}VD, breadth-first search is employed to find the cells $ C_1 $ and $ C_2 $ closest to the start cell $ C_{start} $ and the goal cell $ C_{goal} $ on the GVD, respectively. And the shortest grid paths $ L_1 $ from $ C_{start} $ to $ C_1 $ and $ L_2 $ from $ C_2 $ to $ C_{goal} $ are also computed in this process. Then, the A* search is utilized to search the shortest grid path $ L_3 $ from $ C_1 $ to $ C_2 $ along the GVD edge pixels, in which collision detection is carried out based on the circumscribed radius of the robot. In particular, the cells whose distance to the closest obstacle is less than the circumscribed radius of the robot are considered invalid and eliminated. Therefore, the final searched 2-D grid path consisting of valid cells is guaranteed to be collision-free. As shown by the red path in Fig. \ref{fig:voronoi}, the final shortest Voronoi grid path from $ C_{start} $ to $ C_{goal} $ is composed of $ L_1 $, $ L_3 $, and $ L_2 $.

Based on the shortest Voronoi grid path searched above, a region called Voronoi corridor is constructed as follows. For every cell $ C_k $ in the shortest Voronoi grid path, the minimum distance to the closest obstacle cell $ d_k $ is retrieved from the G\ensuremath{^2}VD. Then, a square bounding box with a side length of $ 2d_k $ is centered on $ C_k $. The obstacle-free cells covered by all bounding boxes make up the Voronoi corridor, as illustrated by the gray region in Fig. \ref{fig:voronoi}. Every time the G\ensuremath{^2}VD updates, the shortest Voronoi grid path is searched from scratch and the Voronoi corridor is constructed based on the newly obtained Voronoi grid path.

Different from the traditional cost map, which is usually constructed by inflating the obstacle cells and typically used to improve the efficiency of collision detection, the Voronoi corridor is constructed to broadly reduce the search space of path searching and improve the search efficiency. Therefore, elaborate collision detection considering the footprint of the robot is not involved when constructing the 2-D Voronoi corridor, which is done in the 3-D path searching stage. In the stage of the Voronoi corridor construction, only the minimum distances from the cells along the Voronoi grid path to the closest obstacle cells are used to make sure that the Voronoi corridor contains the feasible region.

\subsection{Path Searching}
\label{pathsearching}
The shortest Voronoi grid path is far from the actual optimal path, and its piecewise-linear form also does not satisfy the kinodynamic constraints of the robot. To address these issues, a new state lattice-based path planner is proposed to perform fine path searching within the Voronoi corridor.

A typical state lattice-based path planner consists of two parts, namely, graph construction and graph search \cite{likhachev2009planning}. As for graph construction, the 3-D search space $ \left(x, y, \theta\right) $ is discretized first, where $ \left(x, y\right) $ denotes the position of the robot in the world and $ \theta $ represents the heading of the robot. In particular, the orientation space is discretized into 16 angles. Furthermore, the connectivity between two states in the graph is built from motion primitives which fulfill the kinematic constraints of the robot. In this work, a quintic B\'{e}zier curve-based path generation approach described in \cite{zhang2018multilevel} is employed to generate motion primitives from each discretized angle offline. A motion primitive $ \gamma\left(s_1, s_n\right) $ consists of a sequence of internal robot poses $ \left\{ s_1, s_2, \dots, s_n \right\}, s_i = \left(x_i, y_i, \theta_i\right), 1 \le i \le n $ when moving from state $ s_1 $ to state $ s_n $, where $ n $ is the number of poses contained in the motion primitive. As for graph search, the standard A* search is employed, where the 2-D heuristic $ h_{2D} $ proposed in \cite{likhachev2009planning} is utilized to guide the A* search away from those areas with dead-ends. $ h_{2D} $ is constructed by computing the costs of shortest 2-D grid paths from the goal cell to other cells in the search space through dynamic programming.

\begin{figure}[t]
	\centering
	\includegraphics[scale=0.2]{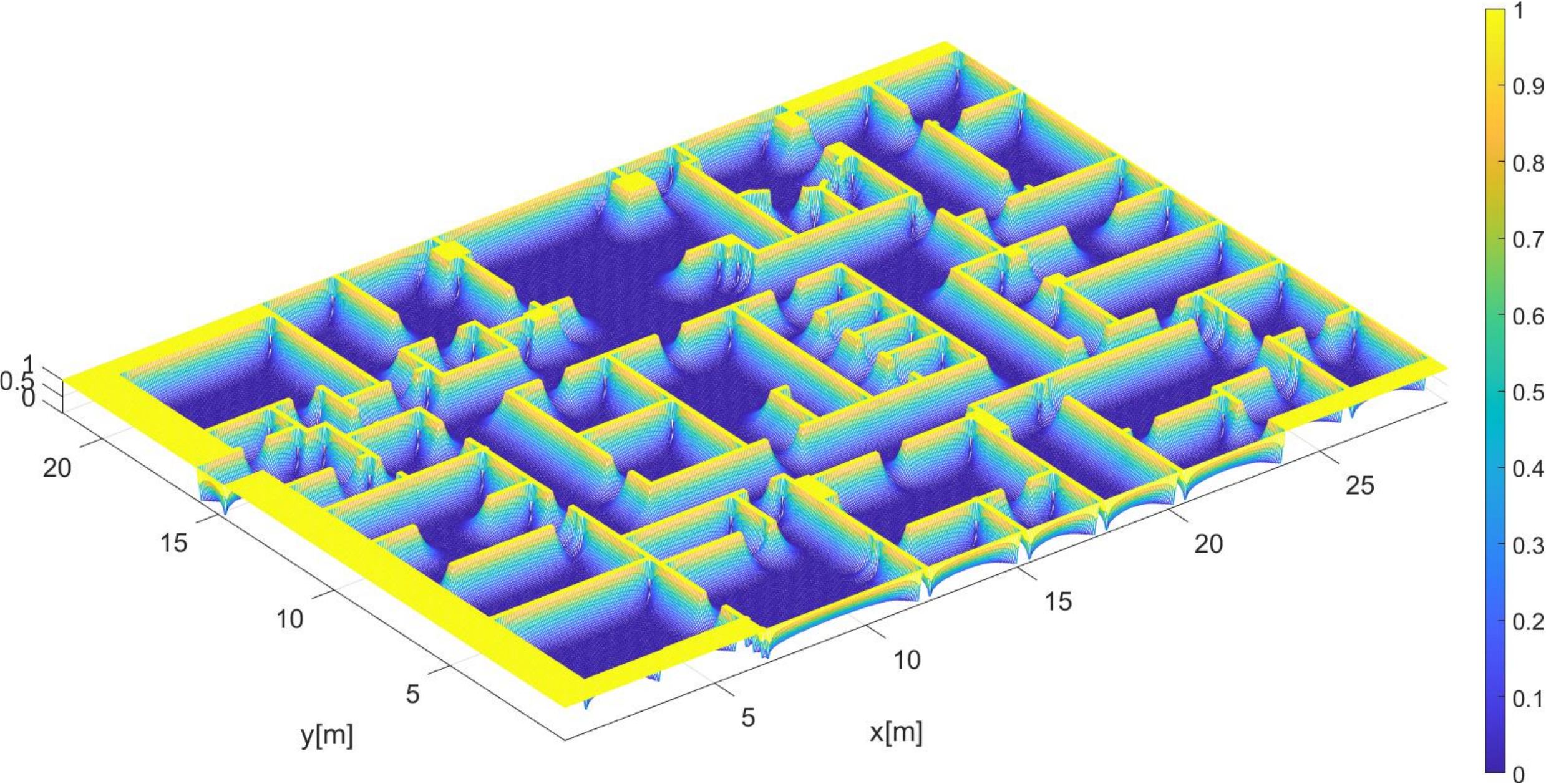}
	\caption{Illustration of the Voronoi field of the office-like environment shown in Fig. \ref{fig:voronoi}. The light yellow regions are obstacle areas.}
	\label{fig:voronoifield}
\end{figure}

In addition, since the motion primitives from each discretized angle are designed offline in advance, the covered cells of each motion primitive considering the robot footprint can be also computed and recorded as a lookup table in advance at the origin $ \left(0, 0\right) $. During every A* expansion, the covered cells of the selected motion primitive are retrieved in the lookup table and translated according to the robot position. As long as one of the covered cells is occupied by obstacles, the corresponding motion primitive is considered invalid and eliminated in this expansion. Therefore, the final searched path consisting of valid motion primitives is guaranteed to be collision-free.

Different from the original state lattice-based path planner \cite{likhachev2009planning} that takes the whole grid map as the search space, we reduce the search space to the Voronoi corridor to further improve the search efficiency. In addition, every time the grid map is updated, $ h_{2D} $ needs recalculation before performing path searching. Since the search space of the newly proposed state lattice-based path planner is reduced to the Voronoi corridor, dynamic programming is accordingly limited to only compute the costs of shortest paths from the goal cell to those cells that are within the Voronoi corridor. Therefore, considerable time is also saved for the recalculation of $ h_{2D} $.

Although the path searched above is optimal in the search space, it may be very close to obstacles since only the path length is considered in the cost function. The ideal path is to keep a reasonable distance from obstacles to provide sufficient optimization margin for subsequent path smoothing. To this end, inspired by the Voronoi field presented in \cite{dolgov2010path}, we design the following potential function
\begin{equation}
	\begin{array}{l}
		{\rho _V}\left( {x,y} \right)\\
		= \left\{ {\begin{aligned}
				&\frac{{{d_{\mathcal{V}}}\left( {x,y} \right)}}{{{d_{\mathcal{O}}}\left( {x,y} \right) + {d_{\mathcal{V}}}\left( {x,y} \right)}}\frac{{{{\left( {{d_{\mathcal{O}}} - d_{\mathcal{O}}^{\min }} \right)}^2}}}{{{{\left( {d_{\mathcal{O}}^{\min }} \right)}^2}}} \ &{{d_{\mathcal{O}}} \le d_{\mathcal{O}}^{\min }}\\
				&0 \ &{{d_{\mathcal{O}}} > d_{\mathcal{O}}^{\min }},\label{eq1}
		\end{aligned}} \right.
	\end{array}
\end{equation}
where $ d_{\mathcal{V}}\left(x, y\right) $ and $ d_{\mathcal{O}}\left(x, y\right) $ denote the distances from the given path vertex $ \left(x, y\right) $ to the closest Voronoi edge and the closest obstacle, respectively. The operation of squaring in Eq. \eqref{eq1} is to make the potential function non-negative. $ d_{\mathcal{O}}^{\min} $ is a threshold specifying the minimum safety distance to obstacles. An example of the Voronoi field is shown in Fig. \ref{fig:voronoifield}.

According to \cite{dolgov2010path}, this potential function has the following properties:
\begin{enumerate}[\hspace{1em}i)]
	\item $ {\rho _V}\left( {x,y} \right) \in [0,1] $ and is continuous on $ \left(x, y\right) $ since we cannot simultaneously have $ d_{\mathcal{V}} = 0 $ and $ d_{\mathcal{O}} = 0 $;
	\item $ {\rho _V}\left( {x,y} \right) $ reaches its maximum only when $ \left(x, y\right) $ is within obstacle areas;
	\item $ {\rho _V}\left( {x,y} \right) $ reaches its minimum only when $ \left(x, y\right) $ is on the edges of the GVD or the distance from $ \left(x, y\right) $ to the closest obstacle is greater than $ d_{\mathcal{O}}^{\min} $.
\end{enumerate} 
It is noteworthy that the potential cost is set to zero when the distance to the closest obstacle is greater than the safety threshold ($ {{d_{\mathcal{O}}} > d_{\mathcal{O}}^{\min }} $). The reason for this design is as follows. Our goal is to make the searched path close to Voronoi edges through the Voronoi field so that the obtained path can keep an appropriate distance from obstacles and provide sufficient optimization margin for subsequent path smoothing. However, if the environment is wider and the Voronoi edge is far from the obstacles on both sides, which may make the searched path far away from the optimal path. Therefore, we set a safety distance threshold $ d_{\mathcal{O}}^{\min } $ and regard the area where the distance to the closest obstacle exceeds $ d_{\mathcal{O}}^{\min } $ as a safe region. In this way, the searched path can keep a certain distance from obstacles and will not be far away from the optimal path through the Voronoi field.

Based on the above Voronoi field, the cost of motion primitives is defined as follows. For the sake of computational efficiency, we broadly follow the work \cite{dolgov2010path} and temporarily assume that the robot travels at constant linear and angular velocities. If a motion primitive $ \gamma\left(s_1, s_n\right) $ collides with obstacles, the cost $ g\left( \gamma\left(s_1, s_n\right) \right) $ is set to infinity. Otherwise, the cost of this motion primitive is defined as
\begin{equation}
	g\left( {\gamma \left( {s_1,s_n} \right)} \right) = t\left( {s_1,s_n} \right) \cdot \left( {\mathop {\max }\limits_{(x,y) \in \mathcal{Q}} {\rho _V}\left( {x,y} \right) + 1} \right),\label{eq3}
\end{equation}
where $ t\left( {s_1,s_n} \right) $ is the minimum travel time spent on $ \gamma\left(s_1, s_n\right) $ assuming uniform motion under the constraints of the maximum linear velocity $ v_{\mathrm{max}} $ and the maximum angular velocity $ \omega_{\mathrm{max}} $
\begin{equation}
	t\left( {{s_1},{s_n}} \right) = \max \left\{ {\frac{{\sum\nolimits_{i = 1}^{n - 1} {\left\| {{s_{i + 1}} - {s_i}} \right\|} }}{{{v_{\max }}}},\frac{{\left| {{\theta _n} - {\theta _1}} \right|}}{{{\omega _{\max }}}}} \right\}.
\end{equation}
$ \mathcal{Q} $ denotes the set of 2-D cells covered by the robot when moving from state $ s_1 $ to state $ s_n $. The ``+1'' operation in Eq. \eqref{eq3} is used to ensure that $ g\left( \gamma\left(s_1, s_n\right) \right) $ is a positive number. Intuitively, the Voronoi field penalizes slightly more those actions for which the robot traverses high potential cost areas (e.g., obstacles) and makes the searched path as close as possible to the Voronoi edges or keep a certain distance from obstacles. In addition, the 2-D heuristic $ h_{2D} $ is also multiplied by the potential costs to be consistent with the cost definition of motion primitives.

\textit{Remark:} In general, the path length is considered in the cost function of path planning approaches. However, the action of rotation in place is contained in the designed motion primitives in this work since our platform is a differential-drive mobile robot. And the cost of this action will be 0 if the cost function of the motion primitive is defined as the path length. Therefore, the travel time spent on the motion primitive is considered in the cost function instead of the path length in this work.

\begin{figure}[t]
	\centering
	\subfigure[Path obtained by the state lattice-based path planner \cite{likhachev2009planning}.]{\includegraphics[scale=0.28]{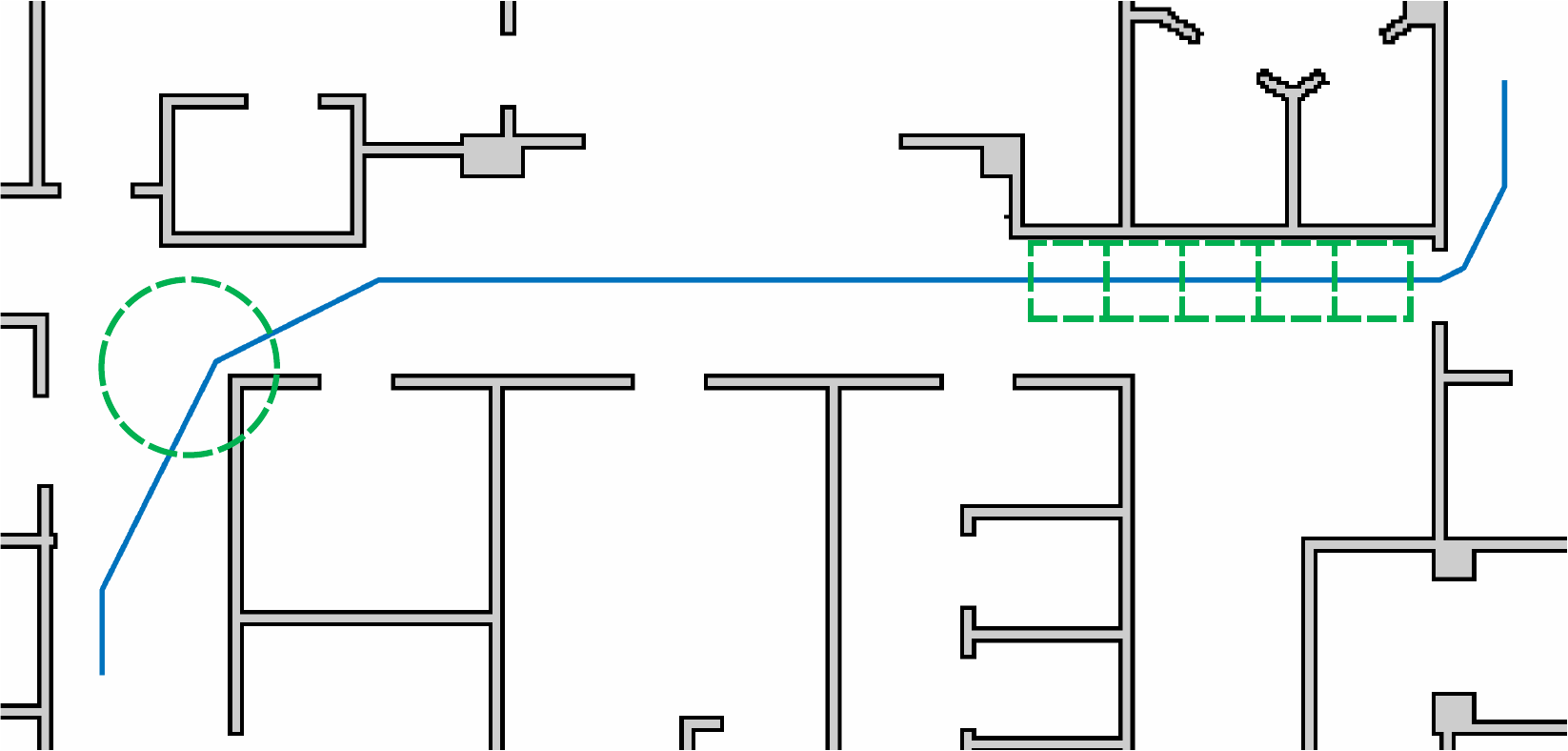}\label{fig:latticepath_1}}
	\centering
	\subfigure[Path obtained by the proposed path planner.]{\includegraphics[scale=0.28]{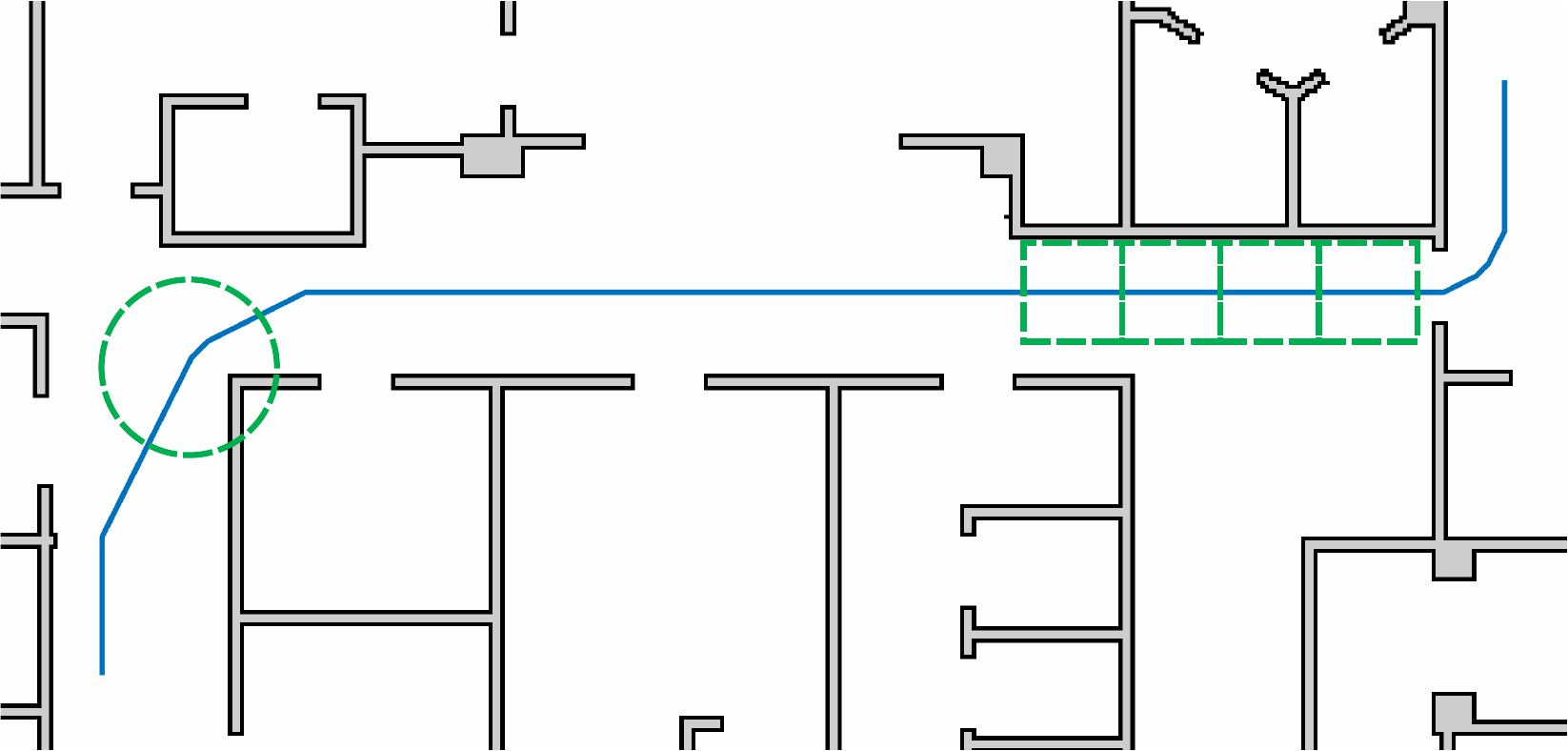}\label{fig:latticepath_2}}
	\caption{Illustration of path searching results.}
\end{figure}

It should be noted that the Voronoi potential field in \cite{dolgov2010path} is used for path smoothing and the region of interest is the whole map. While the search space of the newly proposed state lattice-based path planner is reduced to the Voronoi corridor, and we only need to compute the Voronoi potential costs within the Voronoi corridor. In particular, an efficient distance transform algorithm presented in \cite{felzenszwalb2012distance} is employed to compute the Euclidean distance to the closest Voronoi edge for those cells within the Voronoi corridor. Figs. \ref{fig:latticepath_1} and \ref{fig:latticepath_2} illustrate the paths searched by the original state lattice-based path planner \cite{likhachev2009planning} and the newly proposed path planner, respectively. Compared with the original state lattice-based path planner, the path obtained by the proposed path planner has a certain distance from obstacles and can provide wider optimization margin (green bounding boxes) for further path smoothing, which will be detailed in Section \ref{pathsmoothing}.

\textit{Remark:} The final path is found within the Voronoi corridor, namely, the final solution is searched in a subset of the complete solution set to achieve high computational efficiency performance. The Voronoi corridor is generated based on the shortest Voronoi grid path searched in the G\ensuremath{^2}VD and usually contains the optimal solution in the complete solution set. However, when there is a significantly large difference in the dimension of the environment, it is possible that the optimal solution is contained in a sub-optimal Voronoi corridor derived based on a sub-optimal Voronoi grid path. Therefore, we cannot theoretically guarantee the optimality of the proposed approach. However, the practical performance is satisfactory, which is demonstrated through extensive simulation and experimental tests in Sections \ref{simulations} and \ref{experiments}. In addition, there may be potential extreme scenarios where regional disconnections occur, which will cause the failures of both the Voronoi path search and the Voronoi corridor construction. In these circumstances, the proposed path searching approach degenerates into the original state lattice-based path planner.

\subsection{Path Smoothing}
\label{pathsmoothing}
The path obtained by the state lattice-based path planner is kinematically feasible, but it is still piecewise-linear and not suitable for velocity planning. Therefore, an efficient QP-based path smoothing approach combined with cubic spline interpolation is employed to further smooth the path.

\begin{figure}[t]
	\centering
	\includegraphics[scale=0.48]{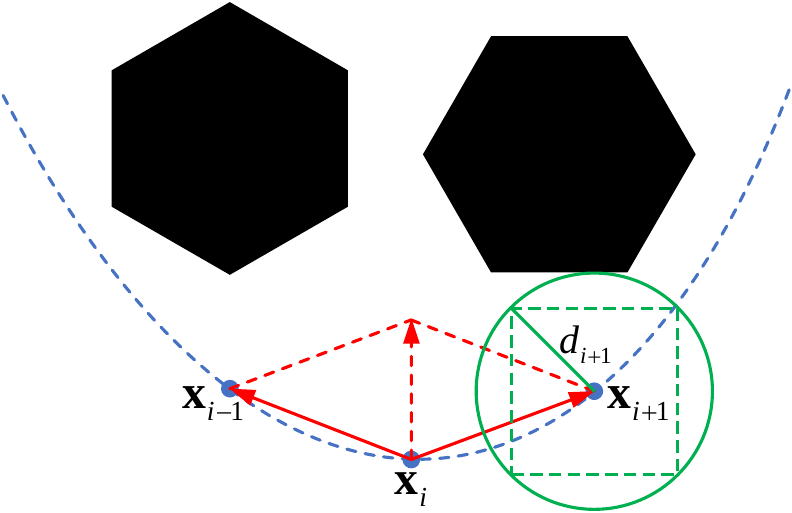}
	\caption{Illustration of path smoothing formulation. $ \mathbf{x}_{i-1} $, $ \mathbf{x}_i $, and $ \mathbf{x}_{i+1} $ denote three successive path vertices. $ d_{i+1} $ is the distance from $ \mathbf{x}_{i+1} $ to the closest obstacle. The black blocks represent obstacles.}
	\label{fig:pathsmoothing}
\end{figure}

The input of path smoothing is several path vertices, which are obtained by sampling in the path generated by the state lattice-based path planner with a fixed interval. Given $n$ reference path vertices to be further smoothed, a convex QP-based path smoothing formulation is defined as
\begin{equation}
	\mathop {\min }\limits_{\mathbf{x}} \, {\omega _s}\sum\limits_{i = 2}^{n - 1} {{{\left\| {{\mathbf{x}_{i + 1}} - 2{\mathbf{x}_i} + {\mathbf{x}_{i - 1}}} \right\|}^2}}  + {\omega _r}\sum\limits_{i = 1}^n {{{\left\| {{\mathbf{x}_i} - {\mathbf{x}_{{i_{ref}}}}} \right\|}^2}},
	\label{eq5}
\end{equation}
subject to
\begin{subequations}
	\begin{align}
		& \mathbf{x}_{1}=\mathbf{x}_{1_{ref}} \text { and } \mathbf{x}_{n}=\mathbf{x}_{n_{ref}}, \\
		& \mathbf{x}_{i} \in \mathcal{B}_{i}, \text { for } i=2, \ldots, n-1, \label{eq6b}
	\end{align}	
\end{subequations}
where $ {\mathbf{x}} = {[{\mathbf{x}}_1^{\mathrm{T}}\;{\mathbf{x}}_2^{\mathrm{T}}\; \ldots \;{\mathbf{x}}_n^{\mathrm{T}}]^{\mathrm{T}}} $ is a $ 2n $-dimensional parameter vector, and $ \mathbf{x}_i = {\left( {x_i}, {y_i} \right)}^{\mathrm{T}}, 1 \le i \le n $ denotes the world coordinates of a path vertex. $ \mathbf{x}_{i_{ref}} = \left(x_{i_{ref}}, y_{i_{ref}}\right)^{\mathrm{T}}, 1 \le i \le n $ is the corresponding reference path vertex of $ \mathbf{x}_i $, and $ \omega_s $ and $ \omega_r $ are the weights of cost terms. $ \mathcal{B}_{i} $ is a state bubble constraining the feasible region of the path vertex $ \mathbf{x}_i $. In this work, $ \mathcal{B}_{i} $ is approximated as an inscribed square of a circle centered on $ \mathbf{x}_{i_{ref}} $, where the radius of the circle is equal to the distance $ d_i $ from $ \mathbf{x}_{i_{ref}} $ to the closest obstacle minus the circumscribed radius $ r_c $ of the robot, as shown in Fig. \ref{fig:pathsmoothing}. The footprint of the robot is approximated as a rectangle in this work. To compute $r_c$, we sequentially calculate the distances from the center of the rectangle to its vertices. And the maximum distance is token as $r_c$. Therefore, $ \mathbf{x}_{i} \in \mathcal{B}_{i} $ in Eq. \eqref{eq6b} is approximated as
\begin{subequations}
	\begin{align}
		& x_{i_{ref}} - b_i \le x_i \le x_{i_{ref}} + b_i, \\
		& y_{i_{ref}} - b_i \le y_i \le y_{i_{ref}} + b_i,
	\end{align}	
	\label{eq7}
\end{subequations}
where the optimization margin $ b_i $ is defined as
\begin{equation}
	b_i = 
	\left\{
	\begin{aligned}
		&\frac{{\sqrt 2 }}{2}{d_i} - {r_c} \ &\frac{{\sqrt 2 }}{2}{d_i} > {r_c}\\
		&0 \ &\frac{{\sqrt 2 }}{2}{d_i} \le {r_c}.
	\end{aligned}
	\right.
	\label{eq8}
\end{equation}
Based on the constraints in Eqs. \eqref{eq7} and \eqref{eq8}, the distance between every optimized path vertex and the corresponding closest obstacle is greater than the circumscribed radius of the robot, thus the final smoothed path is guaranteed to be collision-free.  According to Eq. \eqref{eq8}, a path vertex $ \mathbf{x}_{i} $ will be fixed during the optimization process if the corresponding reference path vertex $ \mathbf{x}_{i_{ref}} $ is close to the obstacle ($ \frac{{\sqrt 2 }}{2}{d_i} \le {r_c} $). Intuitively, a reference path with a certain distance from obstacles can provide sufficient optimization margin for path smoothing. This is one of the main reasons why we introduce the Voronoi field and redesign the cost function of path searching to make the searched path keep a reasonable distance from obstacles in Section \ref{pathsearching}.

The first term in Eq. \eqref{eq5} is a measure of the path smoothness. The cost function $ {{\mathbf{x}_{i + 1}} - 2{\mathbf{x}_i} + {\mathbf{x}_{i - 1}}} $ can be rewritten as $ \left( {{\mathbf{x}_{i + 1}} - {\mathbf{x}_i}} \right) + \left( {{\mathbf{x}_{i - 1}} - {\mathbf{x}_i}} \right) $. As shown in Fig. \ref{fig:pathsmoothing}, from a physical point of view, this cost term treats the path as a series spring system, where $ {{\mathbf{x}_{i + 1}} - {\mathbf{x}_i}} $ and $ {{\mathbf{x}_{i - 1}} - {\mathbf{x}_i}} $ are the forces on the two springs connecting the vertices $ {\mathbf{x}_{i + 1}} $, $ {\mathbf{x}_i} $ and $ {\mathbf{x}_{i - 1}} $, $ {\mathbf{x}_i} $, respectively. If the forces $ {{\mathbf{x}_{i + 1}} - {\mathbf{x}_i}} $ and $ {{\mathbf{x}_{i - 1}} - {\mathbf{x}_i}} $ are equal in size and opposite in direction, the resultant force $ \left( {{\mathbf{x}_{i + 1}} - {\mathbf{x}_i}} \right) + \left( {{\mathbf{x}_{i - 1}} - {\mathbf{x}_i}} \right) $ is a zero vector and the norm is minimum. If all the resultant forces are zero vectors, all the vertices would uniformly distribute in a straight line, and the path is ideally smooth.

The second term in Eq. \eqref{eq5} is used to penalize the deviation from the original safe reference path. As mentioned before, the hard-constrained formulation does not explicitly consider the path clearance, namely, distance from feasible paths to obstacles is ignored. As a result, the optimized path may be close to obstacles. To address this issue, the penalty of the deviation from the original reference path is introduced in the optimization objective. Our goal is to obtain a smooth path while minimizing the deviation between the optimized path and the reference path. Because the reference path searched by the proposed state lattice-based path planner has certain clearance to obstacles, the safety of the path is implicitly considered in the optimization formulation, and the final optimized path will not be close to obstacles. This is also the second reason why we introduce the Voronoi field and redesign the cost function of path searching to improve the path clearance of the searched path, in addition to providing wider optimization margin for path smoothing.

Before path smoothing, the proposed state lattice-based path searching approach is used to find an initial path. Thanks to grid-based planning, the initial path is globally optimal in the sense of resolution. Furthermore, the proposed path smoothing approach formulates the path smoothing problem in the form of convex quadratic programming. The convexity ensures that the obtained solution is the global optimal solution of the underlying path smoothing problem.

Compared with traditional smoothing approaches such as cubic spline interpolation and curve optimization algorithms, the proposed QP-based path smoothing approach densely samples the searched path to serve as optimization variables. Such a framework achieves the maximum control and flexibility of the path shape to deal with complex scenarios, such as U-turns, S-shaped turns, etc.

\subsection{Velocity Profile Generation}
After path optimization, a path that is much smoother than the original reference path is obtained. However, the number of the path vertices is the same as that of the input reference path and the optimized path is still piecewise-linear. Therefore, we further smooth the path via cubic spline interpolation to obtain a continuous curve. Finally, a numerical integration (NI)-based time-optimal velocity planning algorithm presented in \cite{zhang2018multilevel} is employed to generate a feasible linear velocity profile along the smoothed path, i.e., to solve the $t \rightarrow s$ mapping problem mentioned in Section \ref{statement}. The NI-based algorithm can acquire a provably time-optimal trajectory with low computational complexity, which solves the problem by computing the maximum velocity curve (MVC) considering both kinematic and environmental constraints and then performing numerical integration under MVC \cite{shen2017essential,shen2018complete,shen2020real}. Readers can refer to \cite{zhang2018multilevel} for more details about the proofs of feasibility, completeness, and time-optimality of this algorithm.

Finally, the trajectory tracking controller proposed in the textbook \cite{siciliano2009force} is employed to track the desired Cartesian trajectory $ \left\{x_d\left(t\right),y_d\left(t\right),\theta_d\left(t\right),v\left(t\right),\omega\left(t\right) \right\} $ for the differential-drive mobile robot used in this work. In particular, the control law is as
\begin{equation}
	\left\{
	\begin{aligned}
	v&=v_d \cos e_3+k_1 e_1 \\
	\omega&=\omega_d+k_2 v_d \frac{\sin e_3}{e_3} e_2+k_3 e_3,
	\end{aligned}
	\label{eq9}
	\right.
\end{equation}
wherein $ k_1 $, $ k_2 $, and $ k_3 $ are the proportional coefficients, which are set to $ 1.0 $, $ 2.5 $, and $ 2.5 $, respectively. And the error terms of $ e_1 $, $ e_2 $, and $ e_3 $ in Eq. \eqref{eq9} are calculated as follows:
\begin{equation}
	\left[\begin{array}{l}
	e_1 \\
	e_2 \\
	e_3
	\end{array}\right]=\left[\begin{array}{ccc}
	\cos \theta & \sin \theta & 0 \\
	-\sin \theta & \cos \theta & 0 \\
	0 & 0 & 1
	\end{array}\right]\left[\begin{array}{l}
	x_d-x \\
	y_d-y \\
	\theta_d-\theta
	\end{array}\right].
\end{equation}
Readers can refer to \cite{siciliano2009force} for more details about the controller.

\section{Implementation details}
\label{implementationdetails}
\subsection{Setup}
The proposed G\ensuremath{^2}VD planner is implemented in C/C++. The convex QP problem described in Section \ref{pathsmoothing} is solved by an alternating direction method of multipliers (ADMM)-based QP solver, OSQP \cite{osqp}. The reference path vertices of path smoothing are obtained by sampling in the path generated by the state lattice-based path planner with an interval of $0.1$ $\mathrm{m}$. Densely sampling vertices along the path will introduce more optimization variables and increase the computational burden. In this work, the sampling interval is set according to the resolution of the underlying G\ensuremath{^2}VD and the dimension of the robot. The weights $ \omega_s $ and $ \omega_r $ are set to $ 10 $ and $ 1 $, respectively. The minimum safety distance $d_{\mathcal{O}}^{\min }$ is set to $ 0.5 $ $ \mathrm{m} $. The update frequency of the G\ensuremath{^2}VD is set to $20$ $\mathrm{Hz}$. Each update takes an average of $ 10 $ $ \mathrm{ms} $. All the simulations and experiments are tested on a laptop with an Intel Core i7-9750H processor and 16 GB RAM.

\subsection{Metrics}
The proposed Voronoi corridor is integrated into the original state lattice-based path planner (A* + motion primitives) \cite{likhachev2009planning} to derive a new state lattice-based path planner (A* + motion primitives + Voronoi corridor). And the new path planner is compared with the original state lattice-based path planner to validate the effectiveness of the Voronoi corridor. The performance of path searching approaches is evaluated in terms of \emph{computational efficiency} and \emph{memory consumption}. In particular, the number of expanded states and the planning time are used to evaluate the computational efficiency, and the graph size, i.e., the number of created nodes in the search graph, is used to evaluate the memory consumption.

\textit{Remark:} The runtime of the proposed path searching approach includes three parts: searching the shortest Voronoi grid path, constructing the Voronoi corridor, and searching the kinematically feasible path within the Voronoi corridor. In this work, the incrementally updatable GVD construction algorithm \cite{lau2013efficient} is integrated into the mapper module. Therefore, the time used to construct the G\ensuremath{^2}VD is not included in the runtime of the proposed path searching approach.

To validate the effectiveness of the proposed QP-based path smoothing approach, we compare it with an advanced sparse-banded structure-based path smoothing approach SBA \cite{wen2020effmop} and the popular optimization-based timed-elastic-band planner TEB \cite{rosmann2017integrated}. We comprehensively evaluate path smoothing approaches in terms of \emph{computational efficiency}. The weights of the smoothness and safety terms in the objective function of SBA \cite{wen2020effmop} are set to $ 1 $ and $ 10 $ respectively according to the original paper. We believe these parameters were carefully tuned by the authors to achieve the best performance, and using these parameters makes our comparisons more convincing. Similarly, TEB uses the default parameters of its open-source implementation\footnote{\url{https://github.com/rst-tu-dortmund/teb_local_planner}}.

\section{Simulations}
\label{simulations}
In this section, we verify the applicability of the proposed G\ensuremath{^2}VD planner in simulation. The popular Gazebo \cite{koenig2004design} is chosen as the simulator. In this work, we choose the large-scale complex maze scenario\footnote{\url{https://github.com/NKU-MobFly-Robotics/local-planning-benchmark}} designed in \cite{wen2021mrpb} for evaluation. The dimension of the maze scenario is $23.7$ $\mathrm{m} \times 25.5$ $\mathrm{m}$, and the resolution of the grid map is $0.1$ $\mathrm{m/cell}$. To obtain the grid map of the maze scenario, we first build the scenario in Gazebo and then perform a breath-first exploration over the grid starting from the origin of the Gazebo world coordinate system.

\begin{table}[t]
	\centering
	\caption{Quantitative statistics of path planning results in the maze}
	\label{tbl:pathsearching}
	\begin{tabular}{ccccccc}
		\toprule
		\multicolumn{2}{c}{} 	   						& \# of     			& Time     			  	& Graph     			& Path    	\\
		\multicolumn{2}{c}{} 	     					& expands     			& (secs)      		  	& size   				& cost     	\\
		\midrule
		\multirow{2}{*}{Test 1} 		& Lattice  		& 155,715        		&  0.115               	& 161,848        		& 291,590   \\
										& Ours 			& \textbf{140,909}  	&  \textbf{0.097}  	   	& \textbf{144,943}   	& 291,590   \\
		\midrule
		\multirow{2}{*}{Test 2}   		& Lattice 		& 167,912        		&  0.122        	    & 174,356        		& 324,846   \\
										& Ours 			& \textbf{146,300}  	&  \textbf{0.103}  	   	& \textbf{149,912}   	& 324,846   \\
		\midrule
		\multirow{2}{*}{Test 3}   		& Lattice		& 202,680        		&  0.151           		& 210,043      			& 383,376  	\\
										& Ours   		& \textbf{169,949}  	&  \textbf{0.121}  	 	& \textbf{173,716}   	& 383,376   \\
		\bottomrule
	\end{tabular}
\end{table}

\begin{figure}[t]
	\centering
	\subfigure[]{\includegraphics[width=4cm]{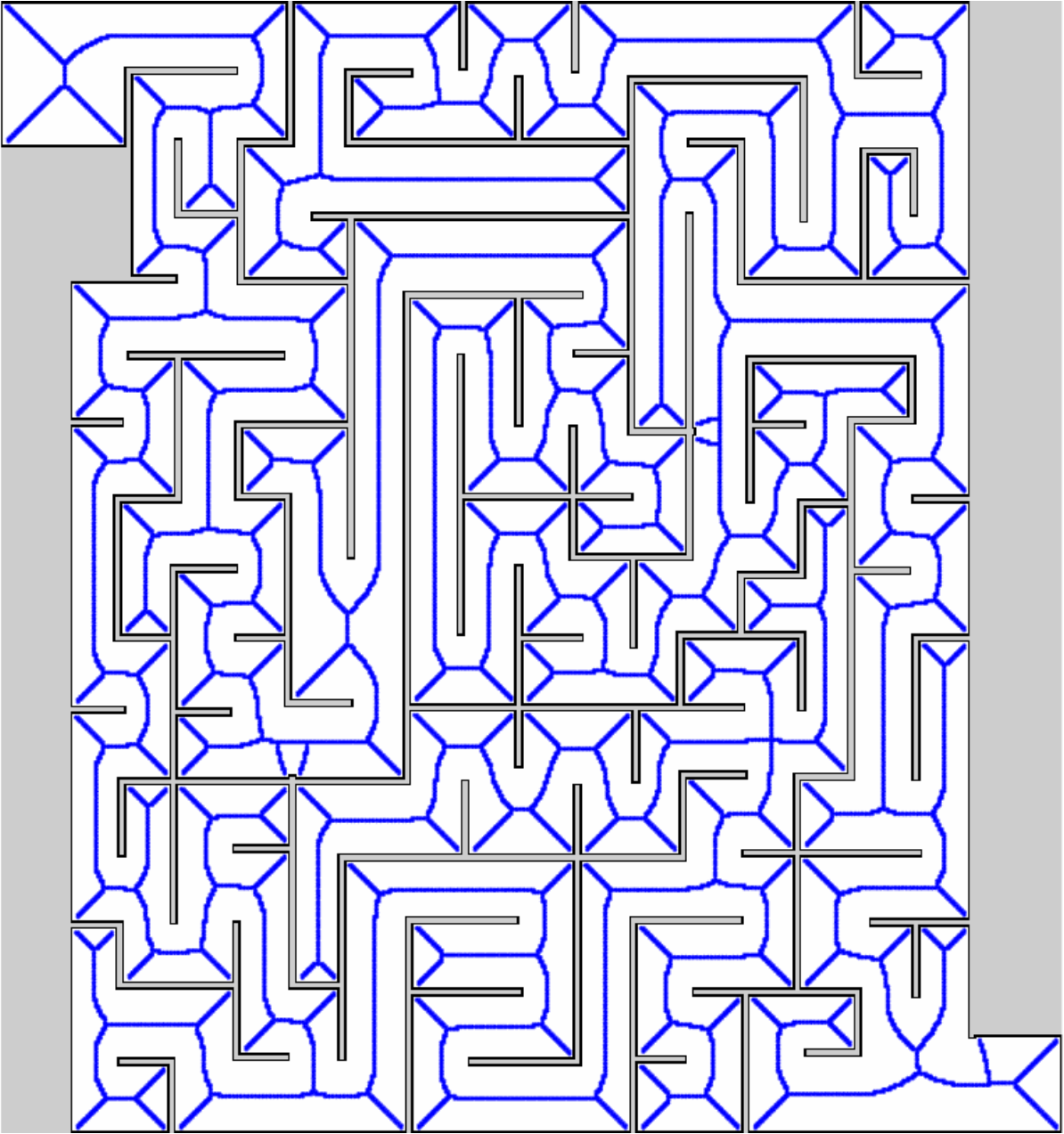}\label{fig:maze_voronoi_edge}}
	\centering
	\subfigure[]{\includegraphics[width=4cm]{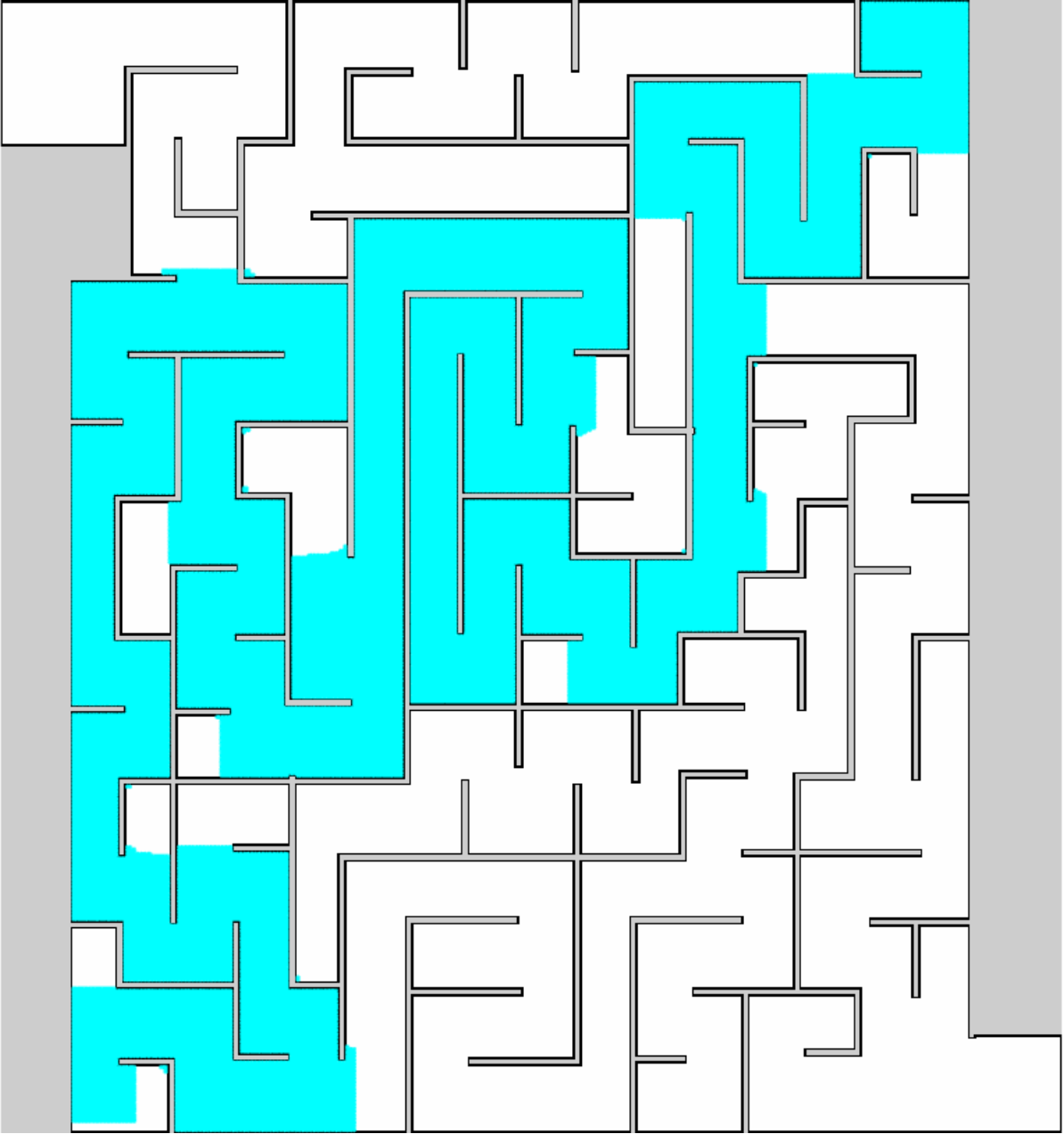}\label{fig:maze_voronoi_corridor}}
	\centering
	\subfigure[]{\includegraphics[width=4cm]{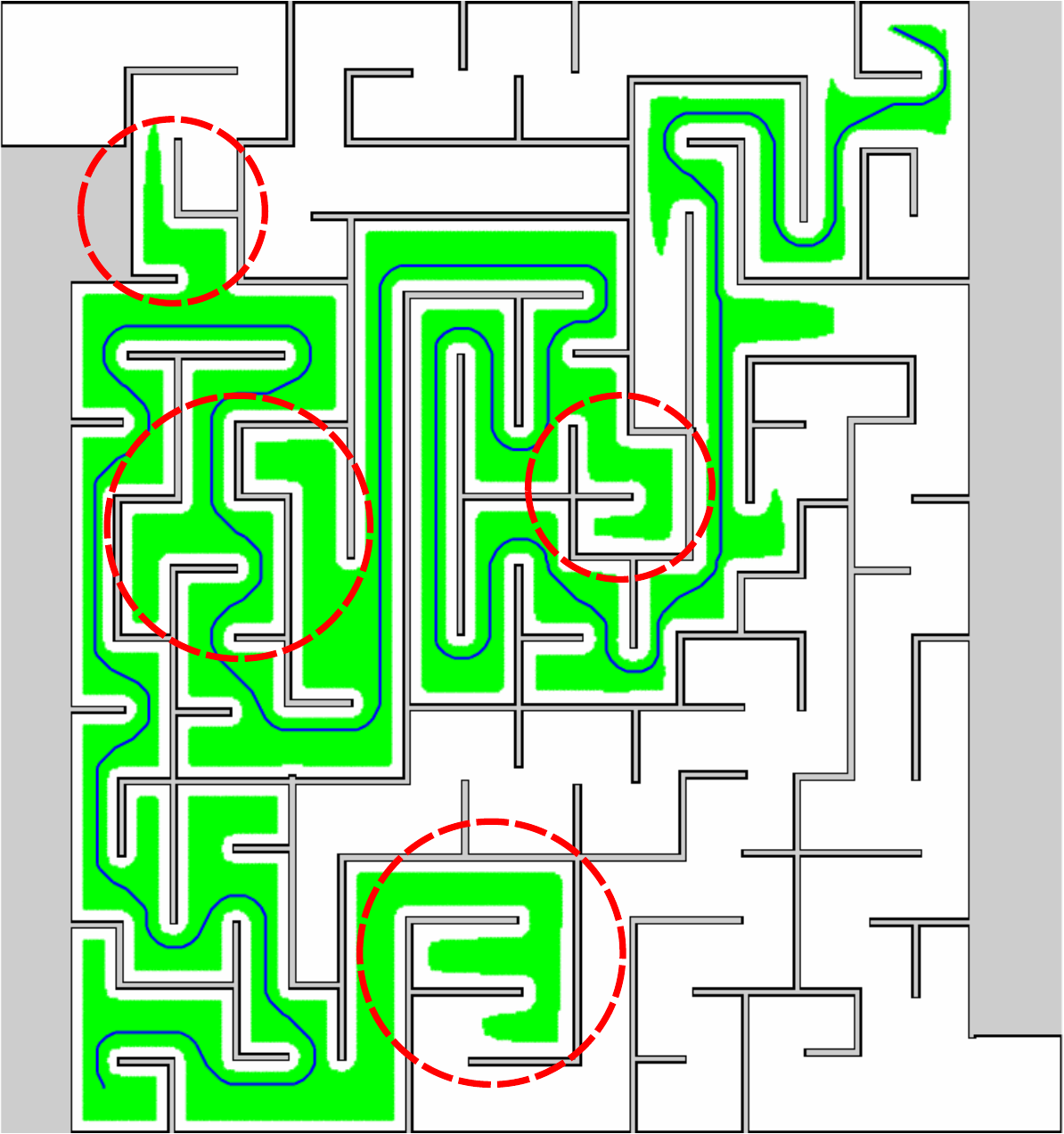}\label{fig:expands_1}}
	\centering
	\subfigure[]{\includegraphics[width=4cm]{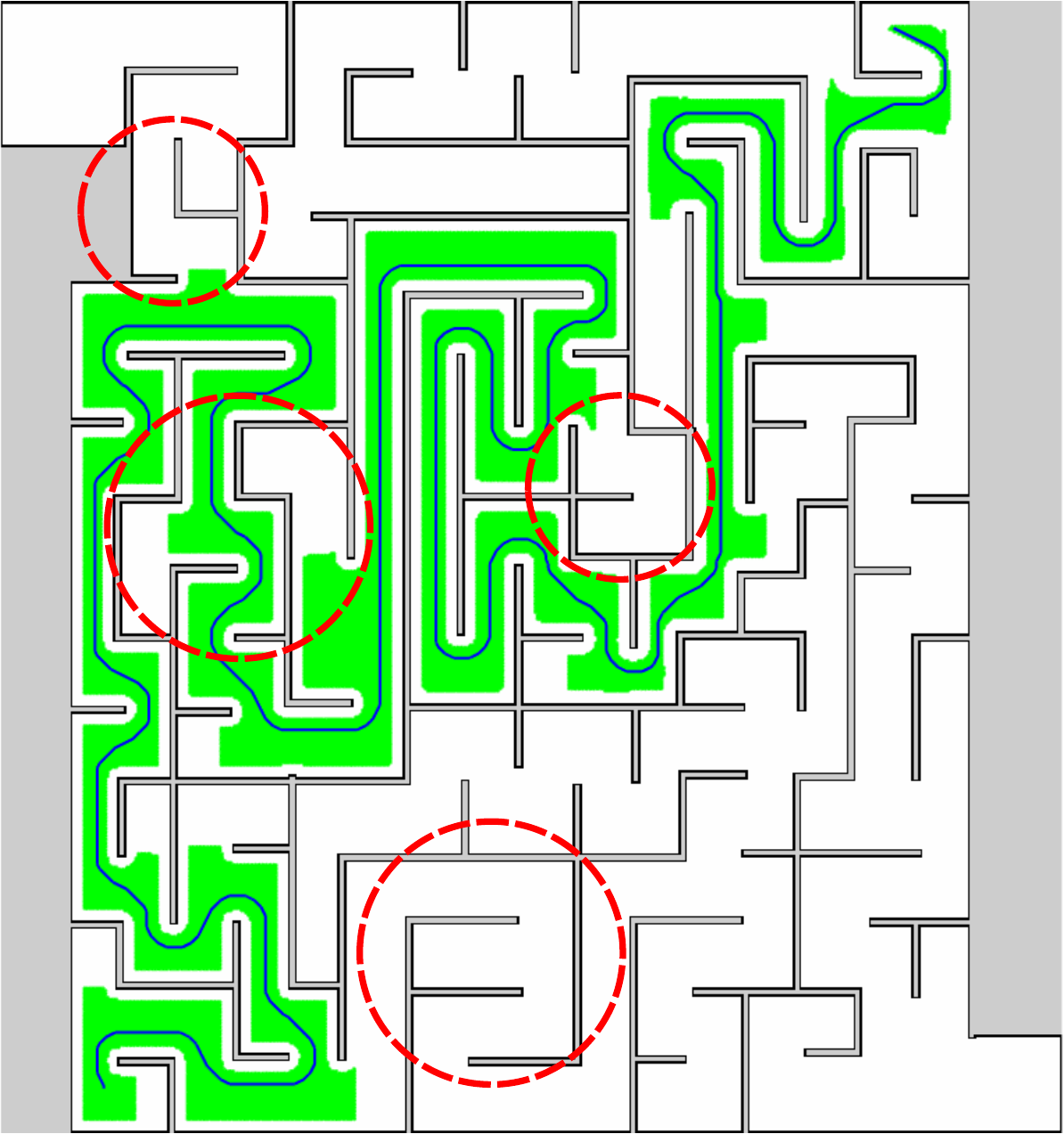}\label{fig:expands_2}}
	\caption{(a) The constructed G\ensuremath{^2}VD. (b) Cyan cells represent the constructed Voronoi corridor. (c) Path planning result of the original state lattice-based path planner. (d) Path planning result of the proposed path planner. The green cells in both (c) and (d) denote the visited cells during the searching process.}
\end{figure}

\begin{figure}[t]
	\centering
	\subfigure[]{\includegraphics[width=4.2cm]{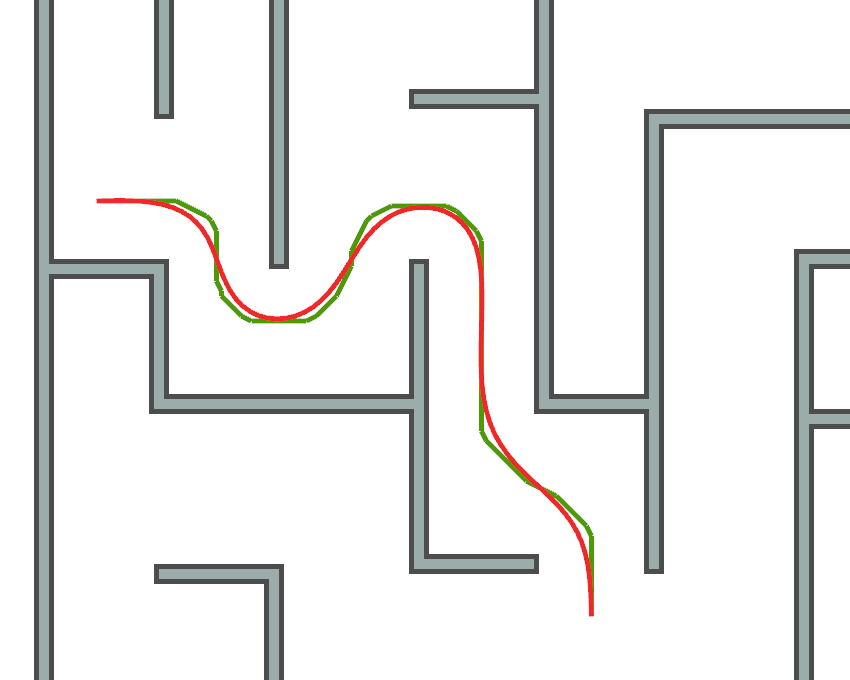}\label{fig:maze_test_1}}
	\hspace{1mm}
	\centering
	\subfigure[]{\includegraphics[width=4.2cm]{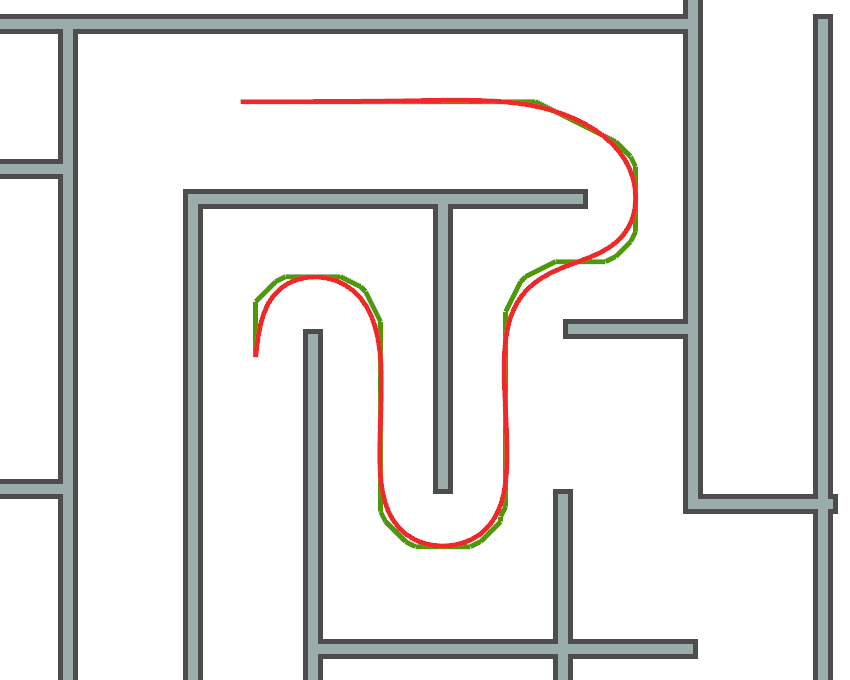}\label{fig:maze_test_2}}
	\centering
	\subfigure[]{\includegraphics[width=4.2cm]{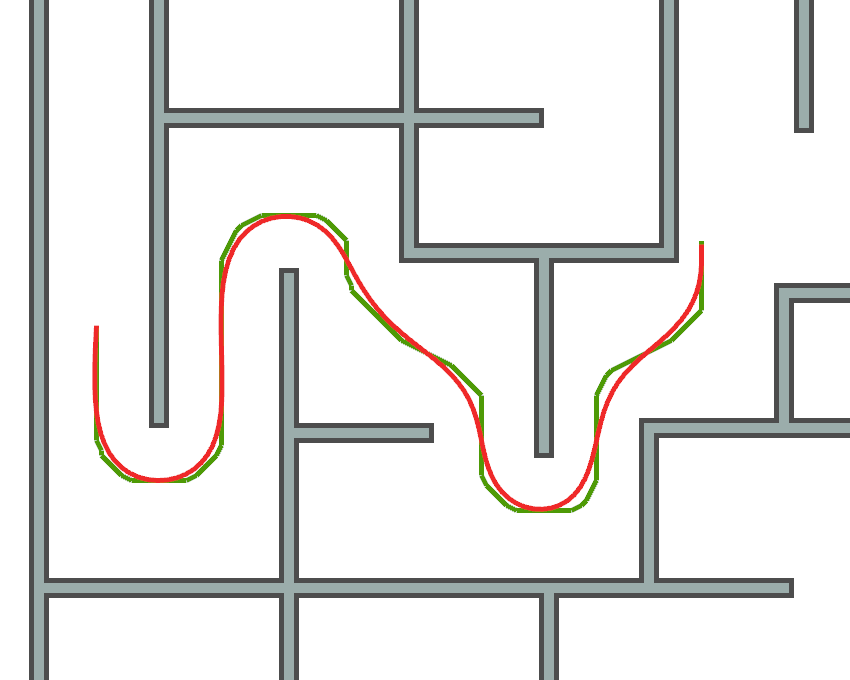}\label{fig:maze_test_3}}
	\hspace{1mm}
	\centering
	\subfigure[]{\includegraphics[width=4.2cm]{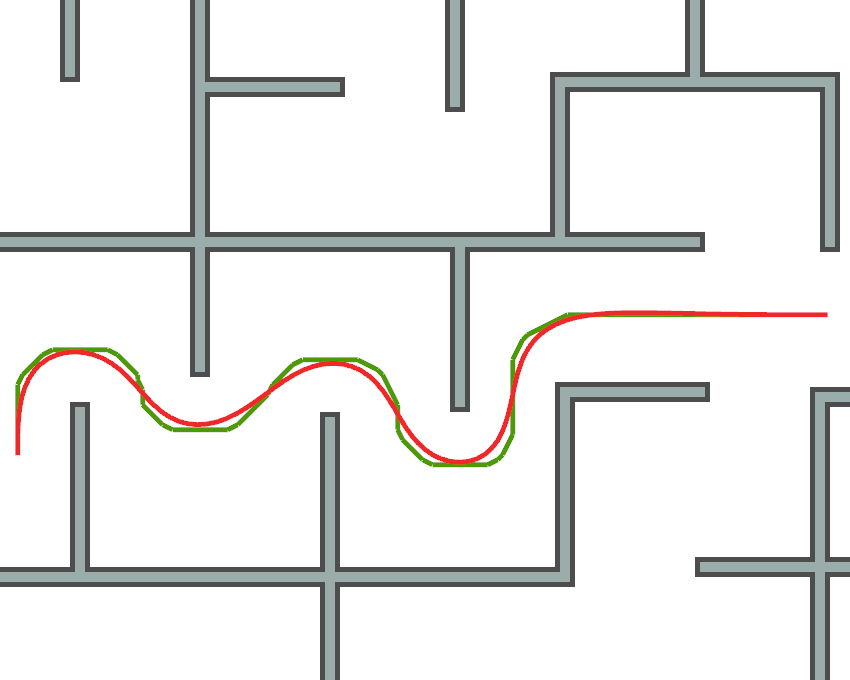}\label{fig:maze_test_4}}
	\caption{Searched paths (green) obtained by the proposed path searching approach and corresponding smoothed paths (red) generated by the proposed path smoothing approach in the maze environment.}
	\label{fig:maze_test}
\end{figure}

\subsection{Comparison on Path Searching}
As mentioned before, the incrementally updatable GVD construction algorithm described in \cite{lau2013efficient} is employed to generate the G\ensuremath{^2}VD of the maze environment, as shown in Fig. \ref{fig:maze_voronoi_edge}. Furthermore, we randomly select three sets of start and goal configurations in the maze for testing. Both the original state lattice-based path planner and the proposed path planner generate equal quality paths. The quantitative statistics of path searching results in these tests are enumerated in Table \ref{tbl:pathsearching}.

The cyan cells in Fig. \ref{fig:maze_voronoi_corridor} represent the constructed Voronoi corridor in Test 1. Figs. \ref{fig:expands_1} and \ref{fig:expands_2} show the path planning results of the two path planners in Test 1, respectively, where the green cells denote the visited cells during the searching process. Compared with the original state lattice-based path planner \cite{likhachev2009planning}, the search space of the proposed path planner is reduced to the Voronoi corridor, thus the search effort spent in those unpromising search areas is significantly saved. As shown in Table \ref{tbl:pathsearching}, the number of expanded search states of the proposed path planner decreases by an average of $ 12.8\% $, and the computational efficiency is improved by $ 17.1\% $. Furthermore, since the search branches for searching in those unpromising search areas are reduced, the number of created nodes in the search graph also decreases. Compared with the original state lattice-based path planner, the graph size of the proposed path planner decreases by an average of $ 13.9\% $.

In conclusion, the newly proposed path planner generates equal quality paths with less time and memory consumption than the original state lattice-based path planner.

\subsection{Comparison on Path Smoothing}

\begin{figure}[t]
	\centering
	\subfigure[Curvature profiles of G\ensuremath{^2}VD, SBA, and TEB in Fig. 7(a).]{
		\includegraphics[height=2.2cm]{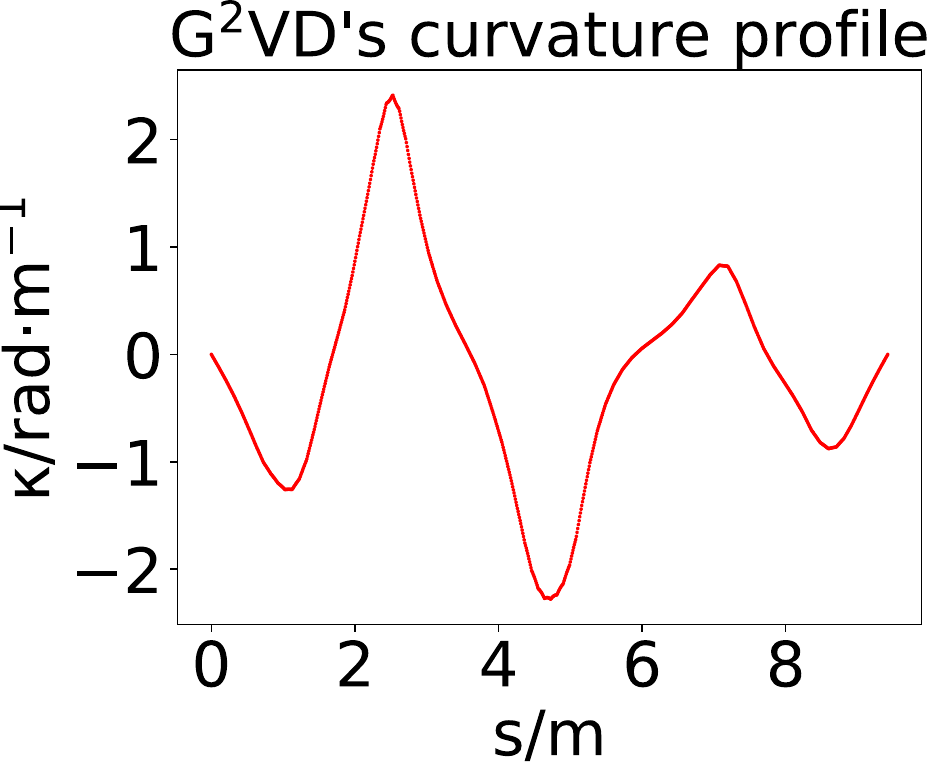}
		\includegraphics[height=2.2cm]{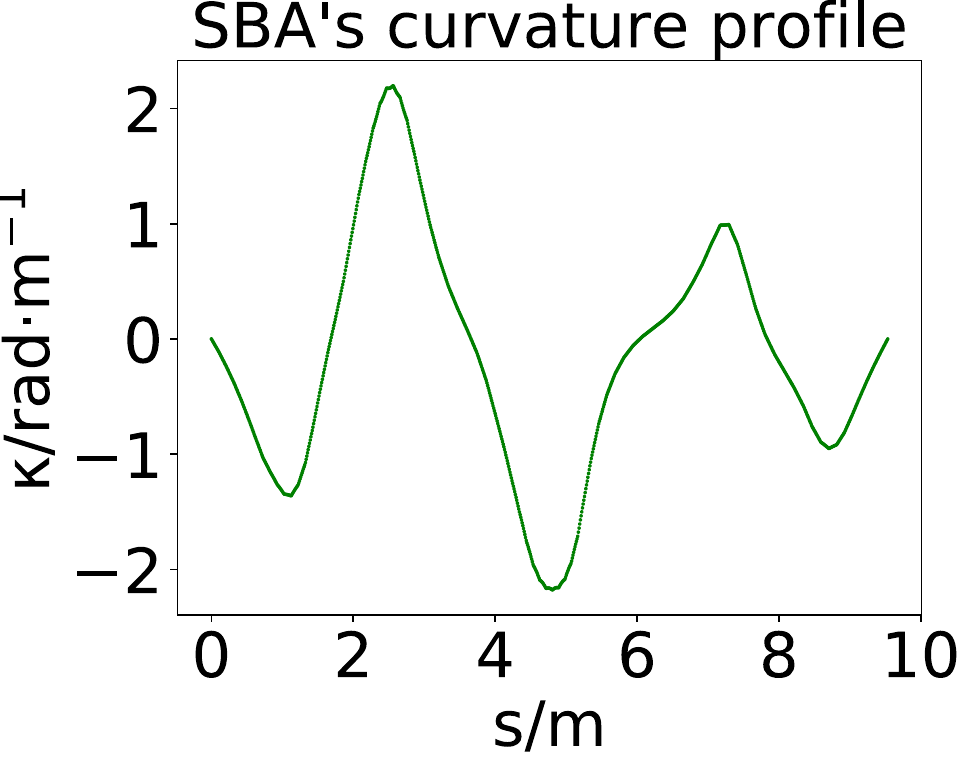}
		\includegraphics[height=2.2cm]{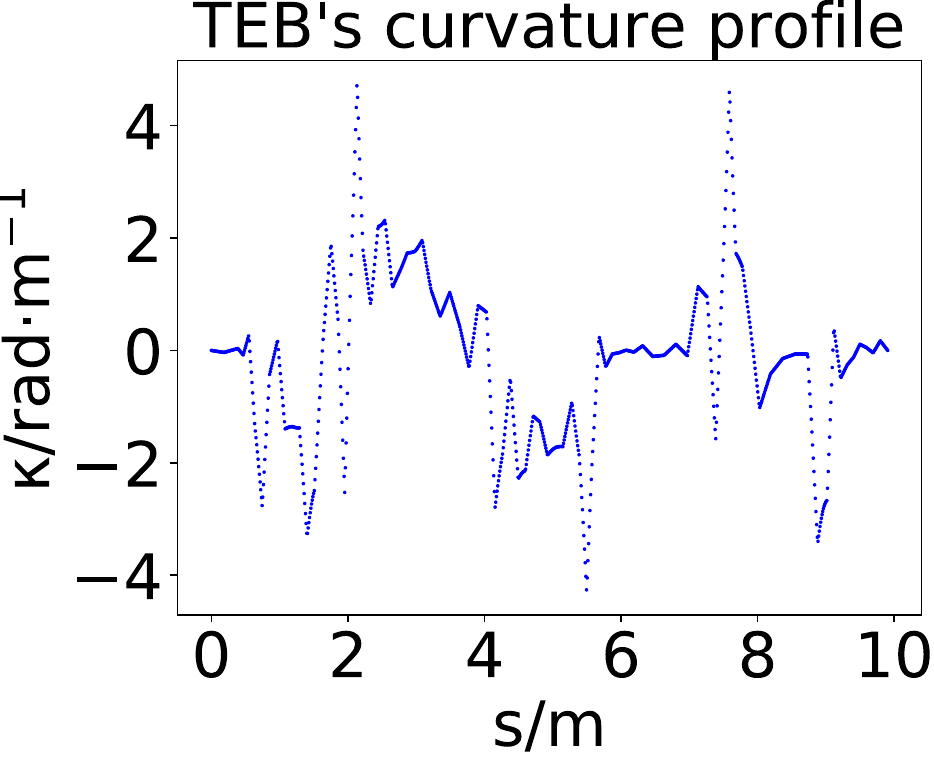}
	}
	\centering
	\subfigure[Curvature profiles of G\ensuremath{^2}VD, SBA, and TEB in Fig. 7(b).]{
		\includegraphics[height=2.15cm]{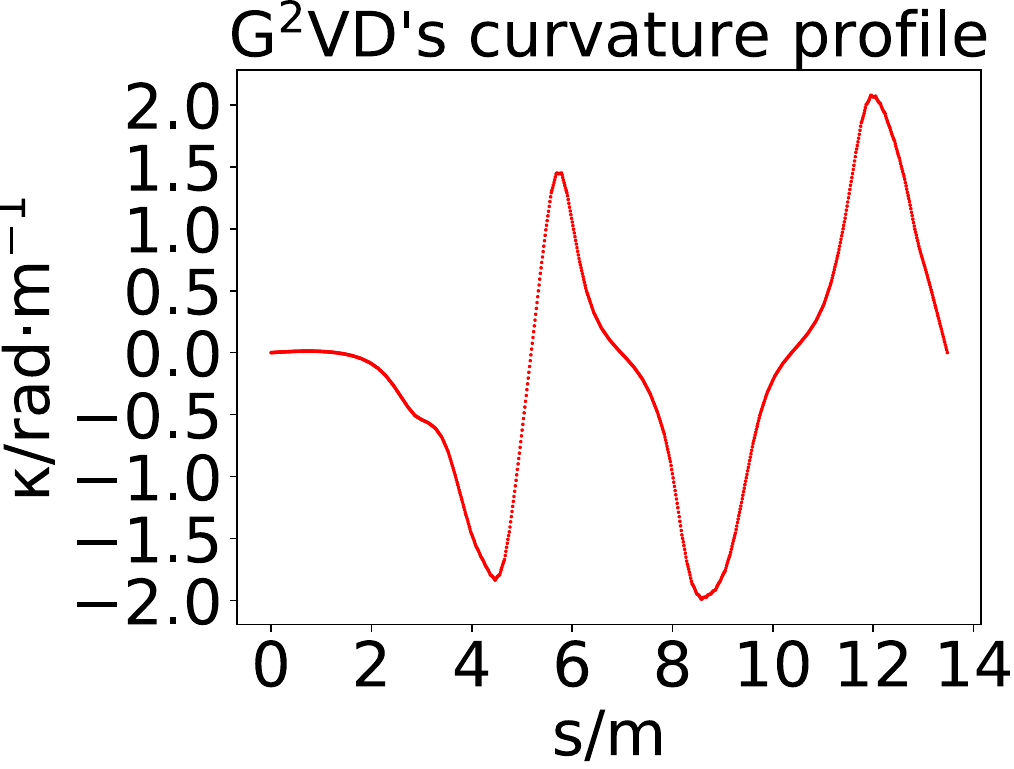}
		\includegraphics[height=2.15cm]{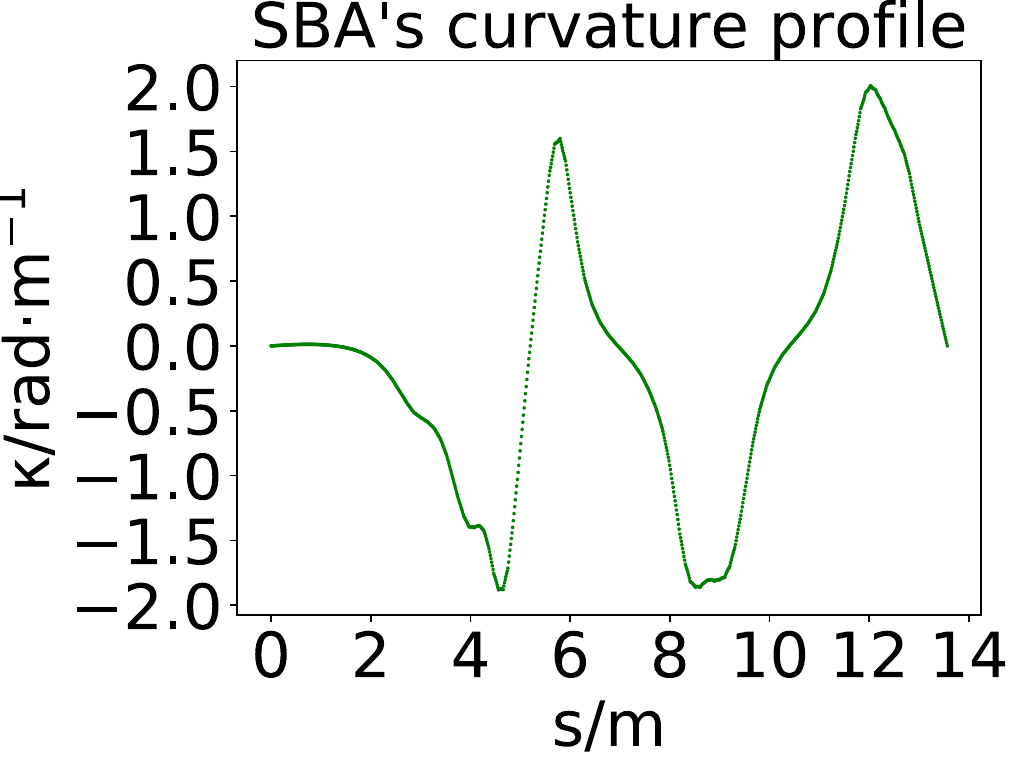}
		\includegraphics[height=2.15cm]{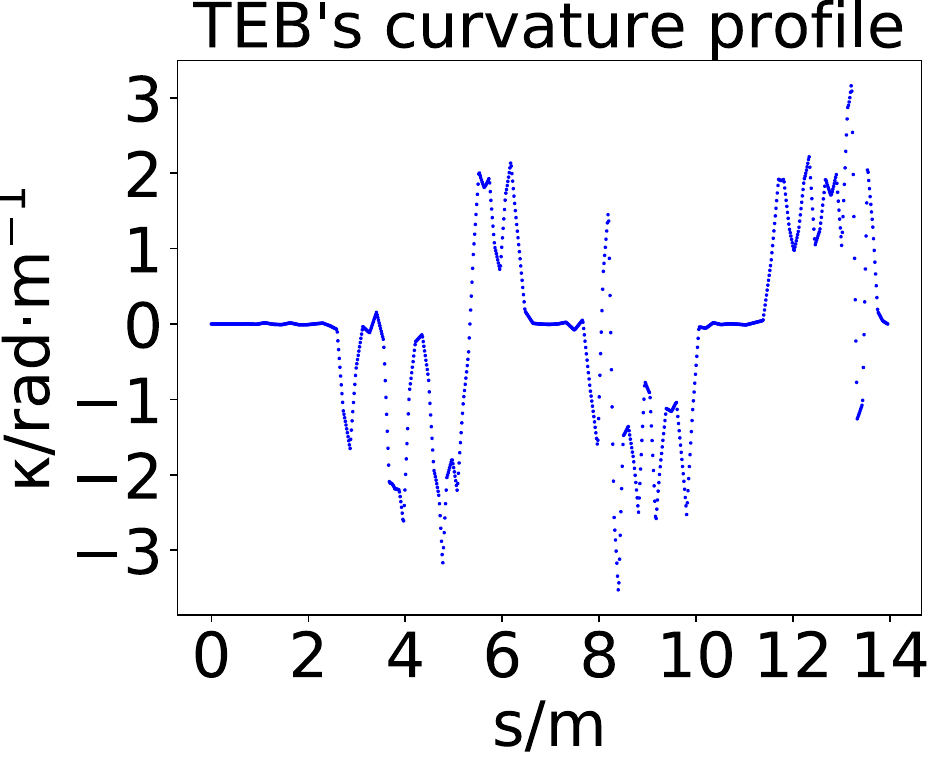}
	}
	\centering
	\subfigure[Curvature profiles of G\ensuremath{^2}VD, SBA, and TEB in Fig. 7(c).]{
		\includegraphics[height=2.2cm]{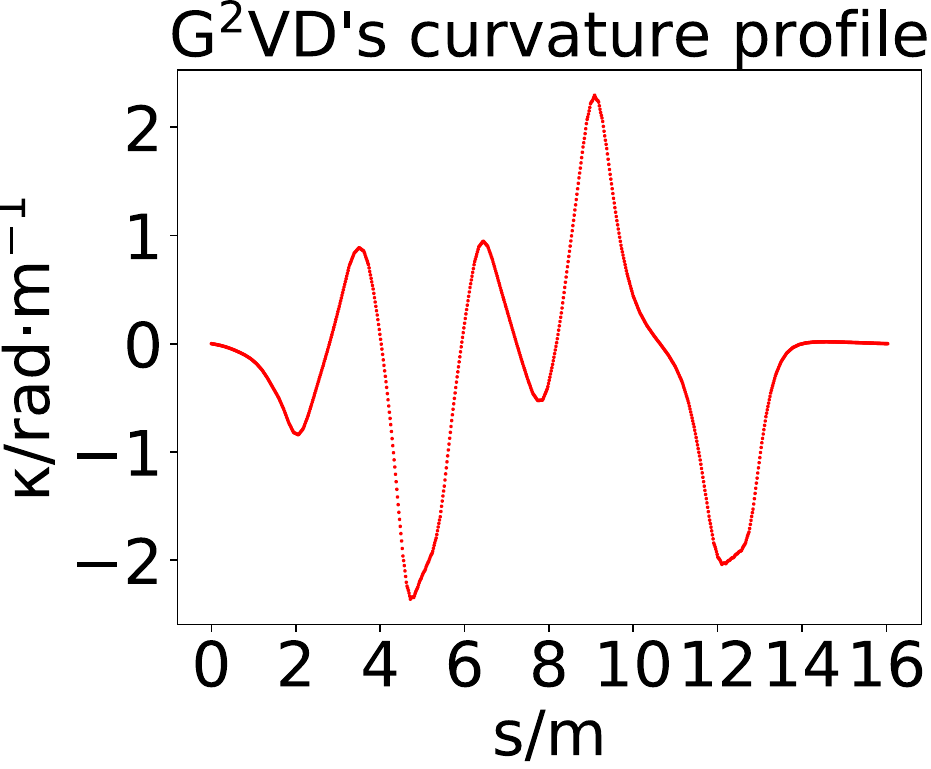}
		\includegraphics[height=2.2cm]{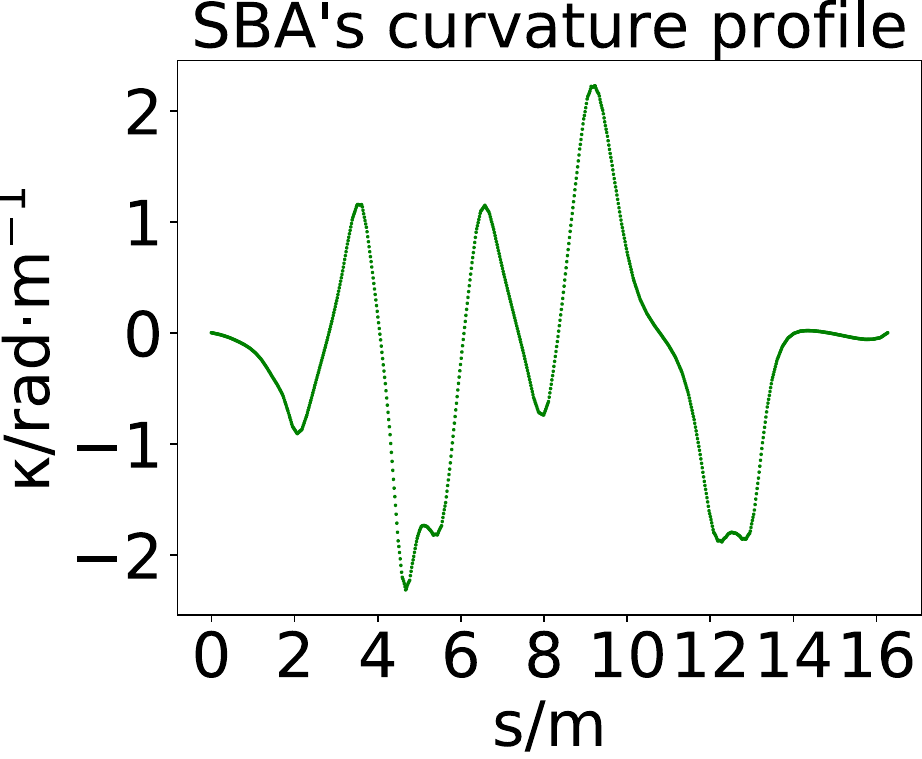}
		\includegraphics[height=2.2cm]{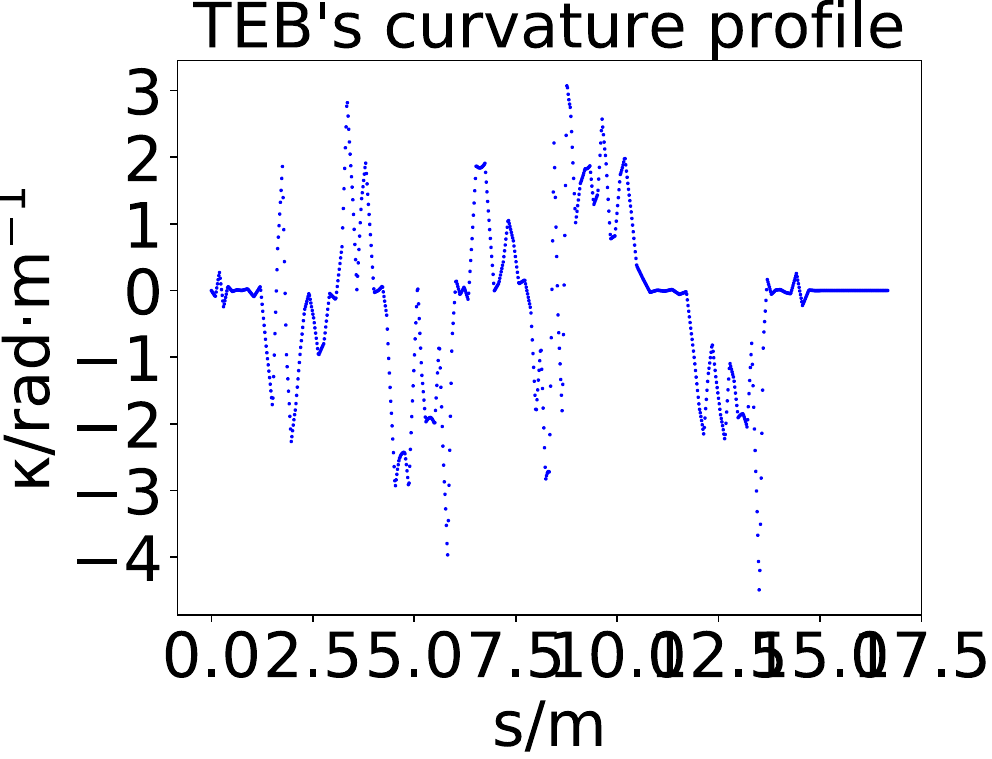}
	}
	\centering
	\subfigure[Curvature profiles of G\ensuremath{^2}VD, SBA, and TEB in Fig. 7(d).]{
		\includegraphics[height=2.2cm]{kappa/gvd_kappa_1.pdf}
		\includegraphics[height=2.2cm]{kappa/sba_kappa_1.pdf}
		\includegraphics[height=2.2cm]{kappa/teb_kappa_1.pdf}
	}
	\caption{Curvature profiles of G\ensuremath{^2}VD, SBA, and TEB in Fig. 7.}
	\label{fig:curvature}
\end{figure}

We choose four challenging local scenarios containing continuous S-shaped or U-shaped turns in the maze environment to compare the computational efficiency of path smoothing approaches. For a fair comparison, the proposed path searching approach is employed to generate the same reference path for SBA \cite{wen2020effmop}, TEB \cite{rosmann2017integrated}, and the proposed QP-based path smoothing approach. In each scenario, we set the same start and goal configurations and repeat the test $ 20 $ times. For testing purposes, we do not limit the length of the initial path for path smoothing, namely, we sample all path points obtained by path searching with an interval of $0.1$ $\mathrm{m}$. The searched paths obtained by the proposed path searching approach and the corresponding smoothed paths generated by the proposed QP-based path smoothing approach are shown in green and red in Fig. \ref{fig:maze_test} respectively, and the statistics of the runtime performance are shown in Table \ref{table:pathsmoothingtime_maze}. According to \cite{wen2020effmop} and \cite{rosmann2017integrated}, both SBA and TEB are soft-constrained path smoothing approaches, wherein both the path smoothness and path clearance to obstacles are considered in the optimization formulation. The terms of path clearance to obstacles are non-convex, resulting in the final optimization formulation is also non-convex. While the proposed path smoothing approach formulates the path smoothing problem in the form of convex quadratic programming, and the convexity allows the problem to be solved efficiently. As shown in Table \ref{table:pathsmoothingtime_maze}, the maximum runtime of the proposed path smoothing approach is less than $ 0.7 $ $ \mathrm{ms} $ in all four tests, and the computational efficiency of the proposed approach is approximatively $ 6.6 $ and $ 53.3 $ times faster than that of SBA and TEB on average, respectively. In addition, the curvature profiles of the paths generated by SBA, TEB, and the proposed path smoothing approach are presented in Fig. \ref{fig:curvature}. Since the smoothness of the path is explicitly considered in the optimization objective of our approach and SBA, their curvature profiles are much smoother than that of TEB.

\begin{table}[t]
	\centering
	\caption{Computational time (in millisecond) of path smoothing in the maze}
	\label{tbl:pathsmoothingtime}
	\begin{tabular}{cccccc}
		\toprule
		\multicolumn{2}{c}{} 	   									& Mean     			& Max     			& Min     				& Std    			\\
		\midrule
		\multirow{3}{*}{Fig. \ref{fig:maze_test_1}} 	& TEB  		& 31.65				& 33.96        		&  29.51                & 1.16   			\\
														& SBA  		& 3.55				& 3.61        		&  3.46                 & 0.04   			\\
														& Ours 		& \textbf{0.49}		& \textbf{0.50}  	&  \textbf{0.48}  	   	& \textbf{0.01}   	\\
		\midrule
		\multirow{3}{*}{Fig. \ref{fig:maze_test_2}}   	& TEB 		& 32.26				& 33.90     		&  29.50        	    & 1.31   			\\
														& SBA 		& 4.24				& 4.29        		&  4.20        	        & 0.03   			\\
														& Ours 		& \textbf{0.61}		& \textbf{0.67}  	&  \textbf{0.60}  	   	& \textbf{0.02}   	\\
		\midrule
		\multirow{3}{*}{Fig. \ref{fig:maze_test_3}}   	& TEB		& 34.65				& 36.23        		&  32.64           	    & 1.02  			\\
														& SBA		& 5.57				& 5.62        		&  5.53           	    & 0.02  			\\
														& Ours   	& \textbf{0.64}		& \textbf{0.65}  	&  \textbf{0.64}  	   	& \textbf{0.01}   	\\
		\midrule
		\multirow{3}{*}{Fig. \ref{fig:maze_test_4}}   	& TEB		& 37.38				& 38.67        		&  35.05           	    & 1.27  			\\
														& SBA		& 3.27				& 3.63        		&  3.22           	    & 0.08  			\\
														& Ours   	& \textbf{0.51}		& \textbf{0.55}  	&  \textbf{0.49}  	   	& \textbf{0.01}   	\\
		\bottomrule
	\end{tabular}
	\label{table:pathsmoothingtime_maze}
\end{table}

In conclusion, the newly proposed QP-based path smoothing approach achieves a significant performance improvement in terms of computational efficiency.

\begin{figure}[t]
	\centering
	\includegraphics[width=7cm]{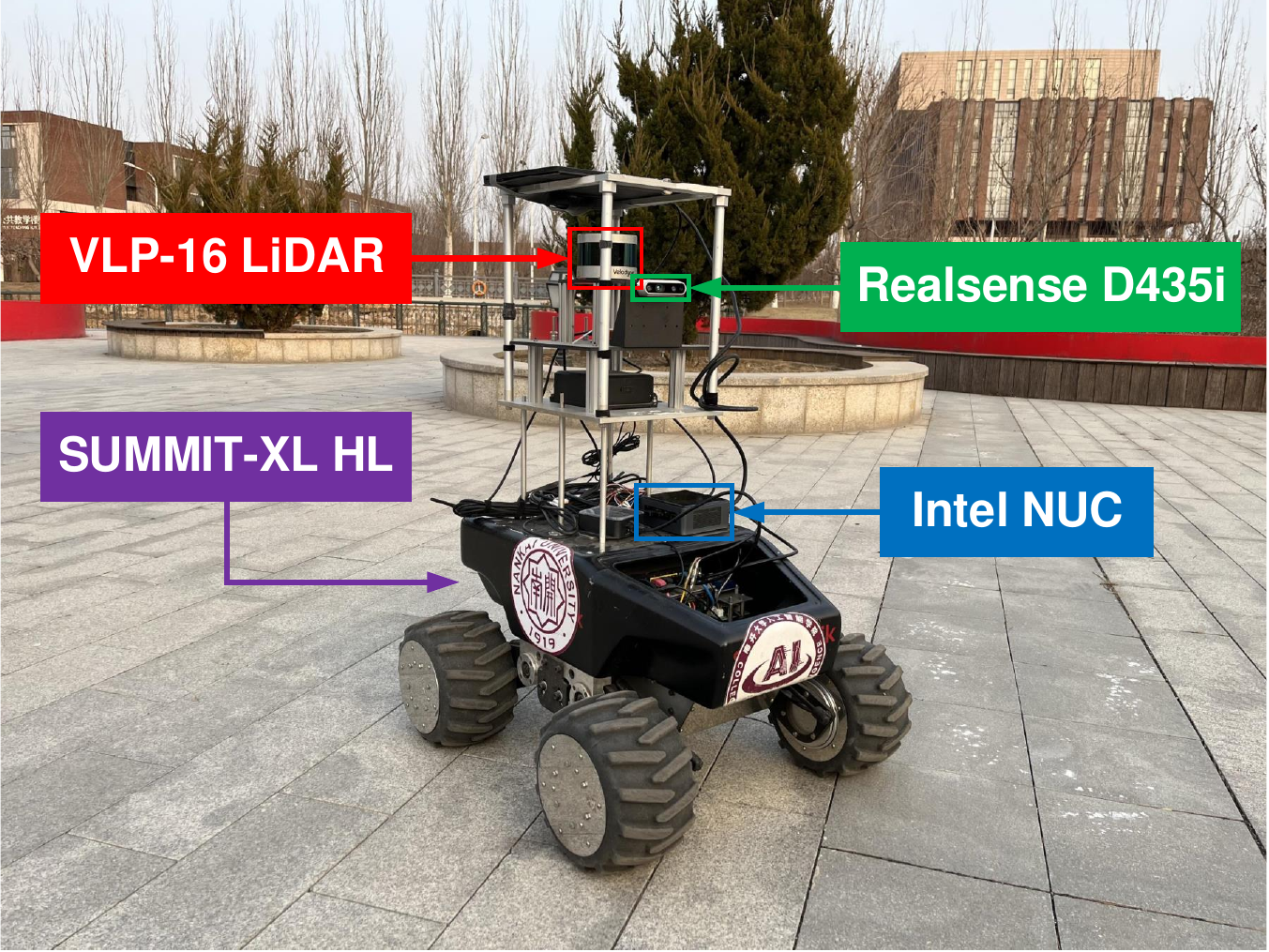}
	\caption{Experimental platform. Our experimental mobile robot SUMMIT-XL HL is equipped with a Velodyne VLP-16 LiDAR, an Intel Realsense D435i depth camera, and an Intel NUC computer.}
	\label{fig:platform}
\end{figure}

\section{Experiments}
\label{experiments}
In this section, outdoor experimental results on our campus are presented and analyzed to validate the effectiveness of the proposed G\ensuremath{^2}VD planner.

\begin{figure}[t]
	\centering
	\includegraphics[scale=0.2]{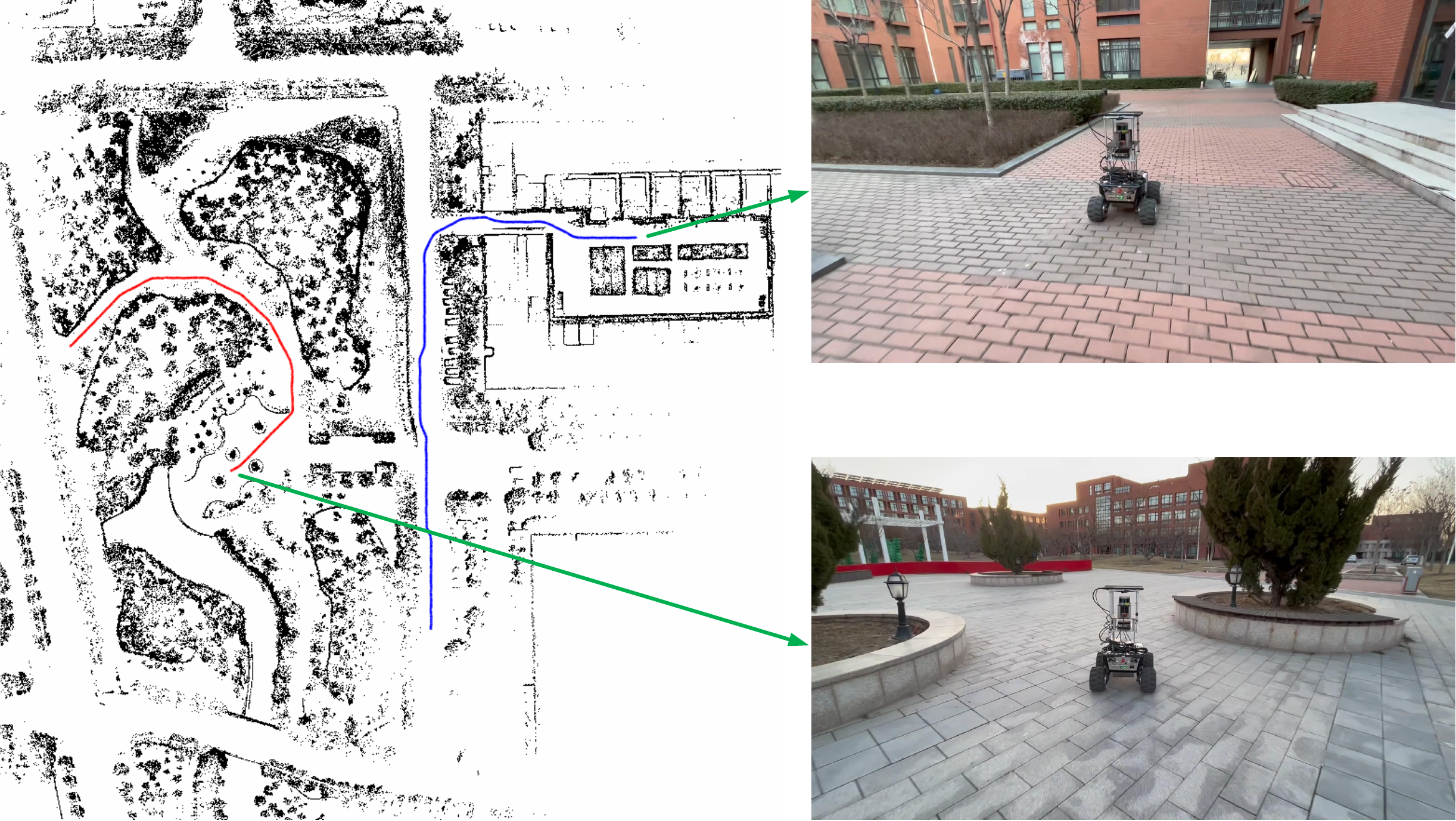}
	\caption{Occupancy grid map for outdoor navigation. The dimension of the environment is approximately $ 210 $ $ \mathrm{m} \times 220 $ $ \mathrm{m} $. The routes of two sets of robot navigation are colored in red and blue, respectively.}
	\label{fig:outdoor}
\end{figure}

\begin{figure}[t]
	\centering
	\subfigure[]{
		\includegraphics[width=4cm]{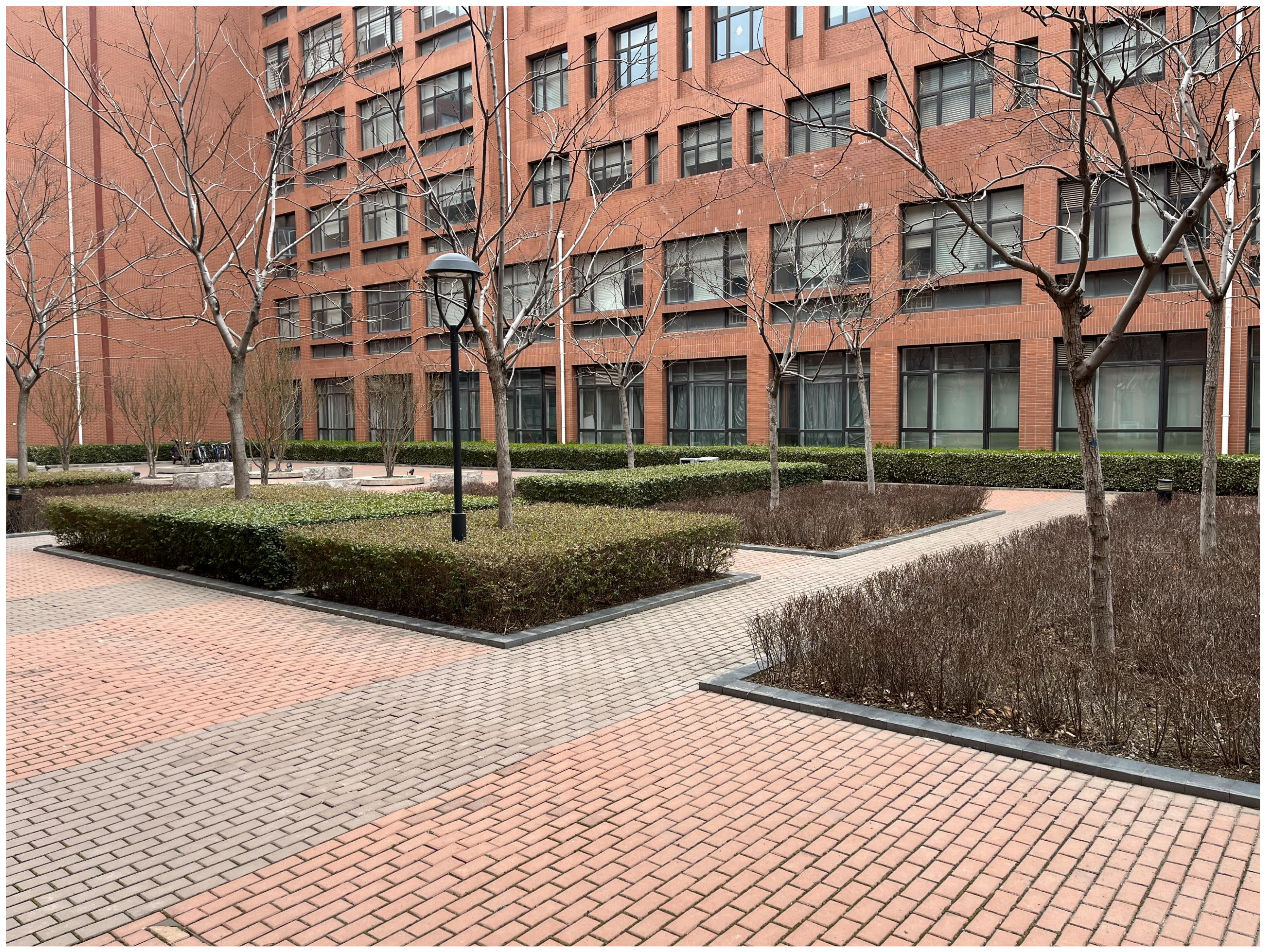}
		\includegraphics[width=4cm]{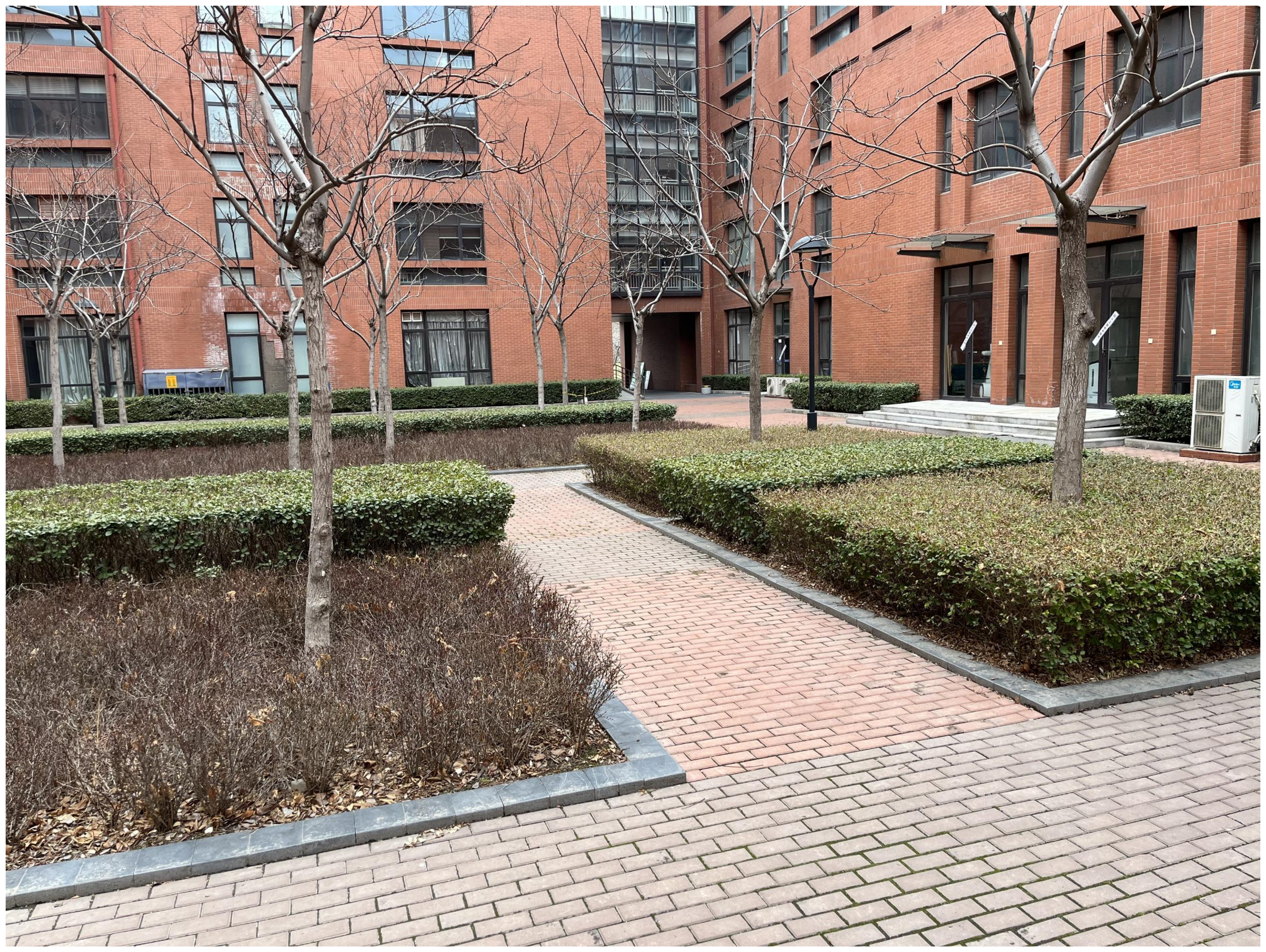}
		\label{fig:outdoor_env}
	}
	\centering
	\subfigure[]{\includegraphics[width=4cm]{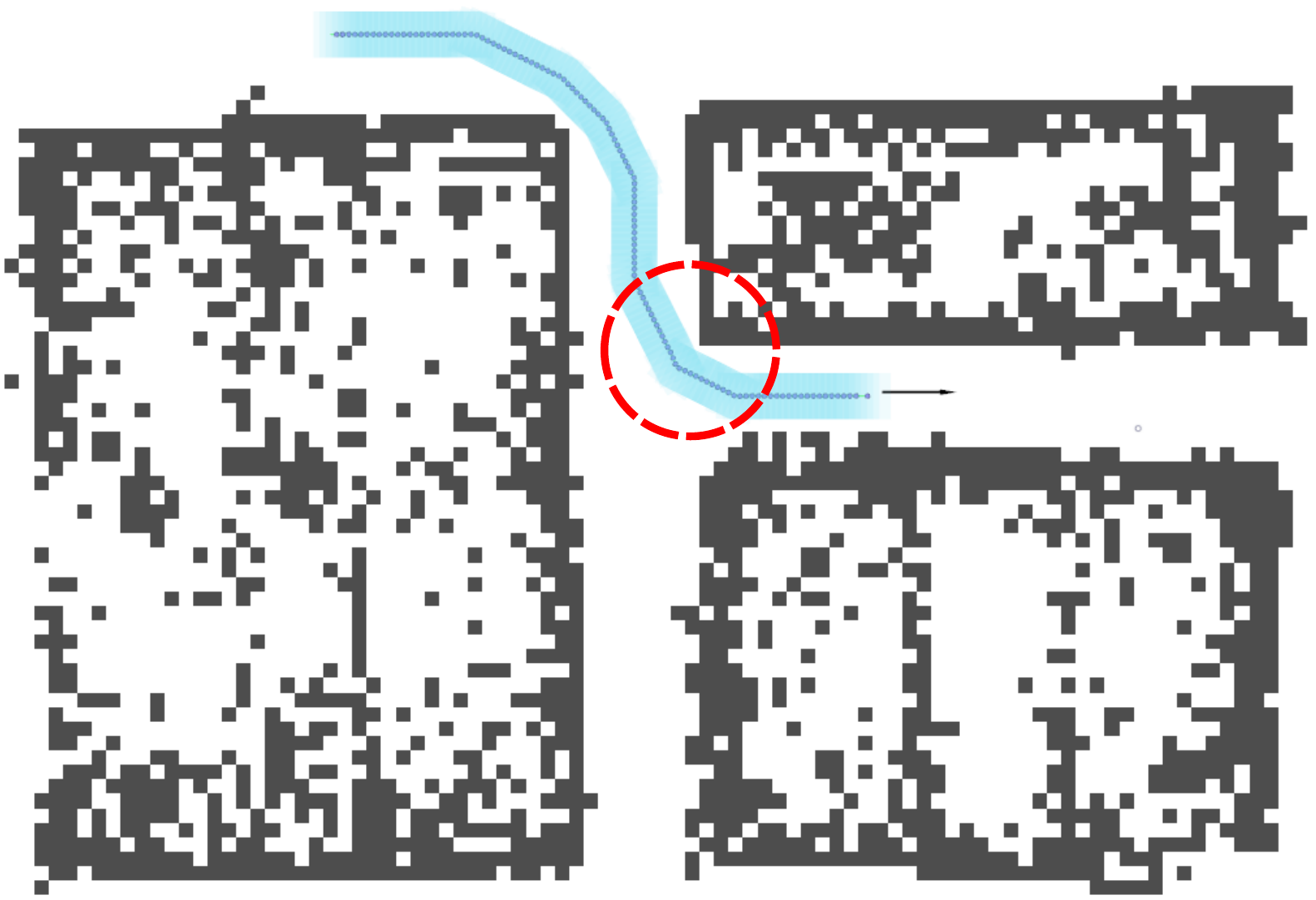}\label{fig:outdoor_lattice_path_1}}
	\centering
	\subfigure[]{\includegraphics[width=4cm]{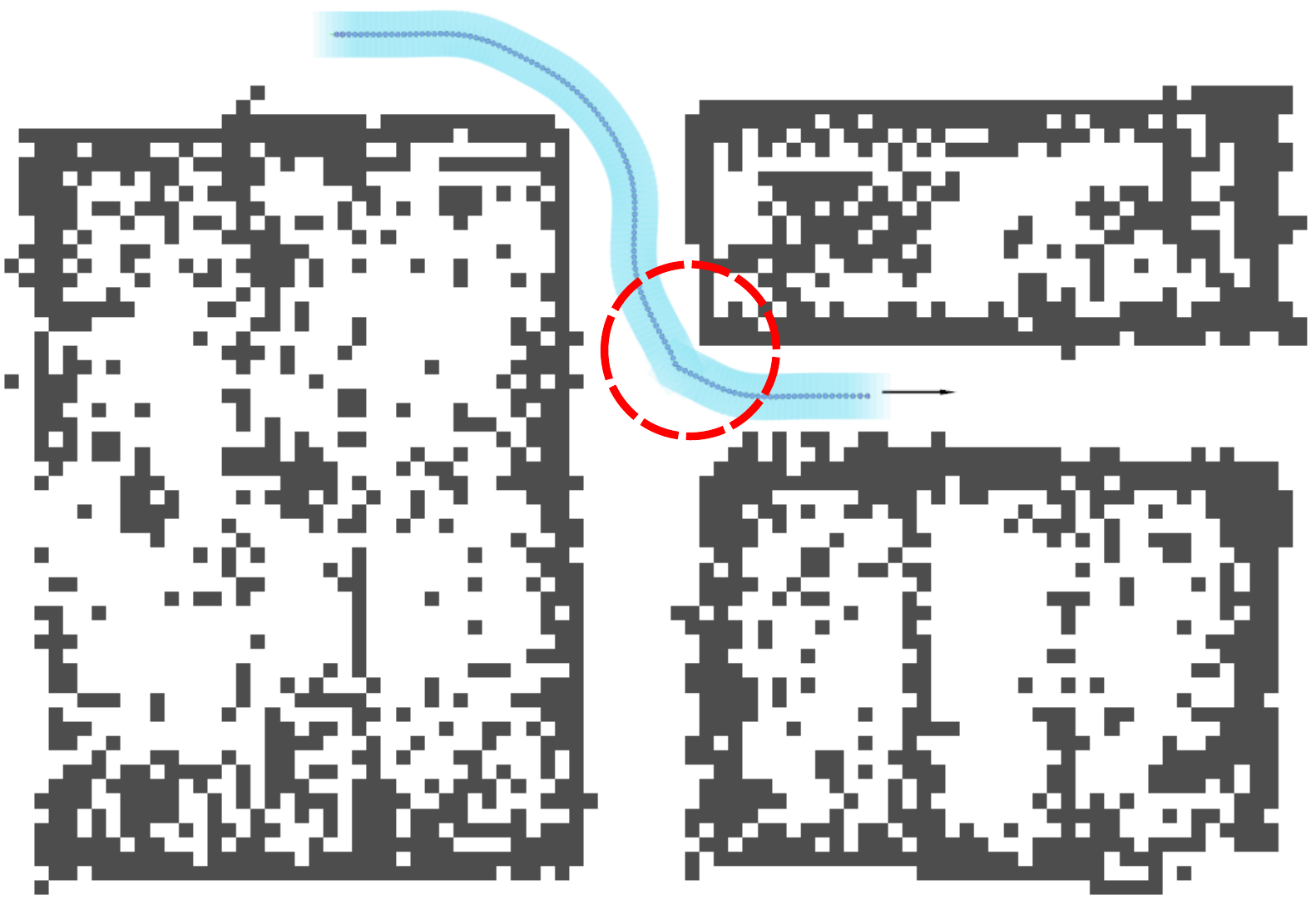}\label{fig:outdoor_lattice_path_2}}
	\subfigure[]{\includegraphics[width=4cm]{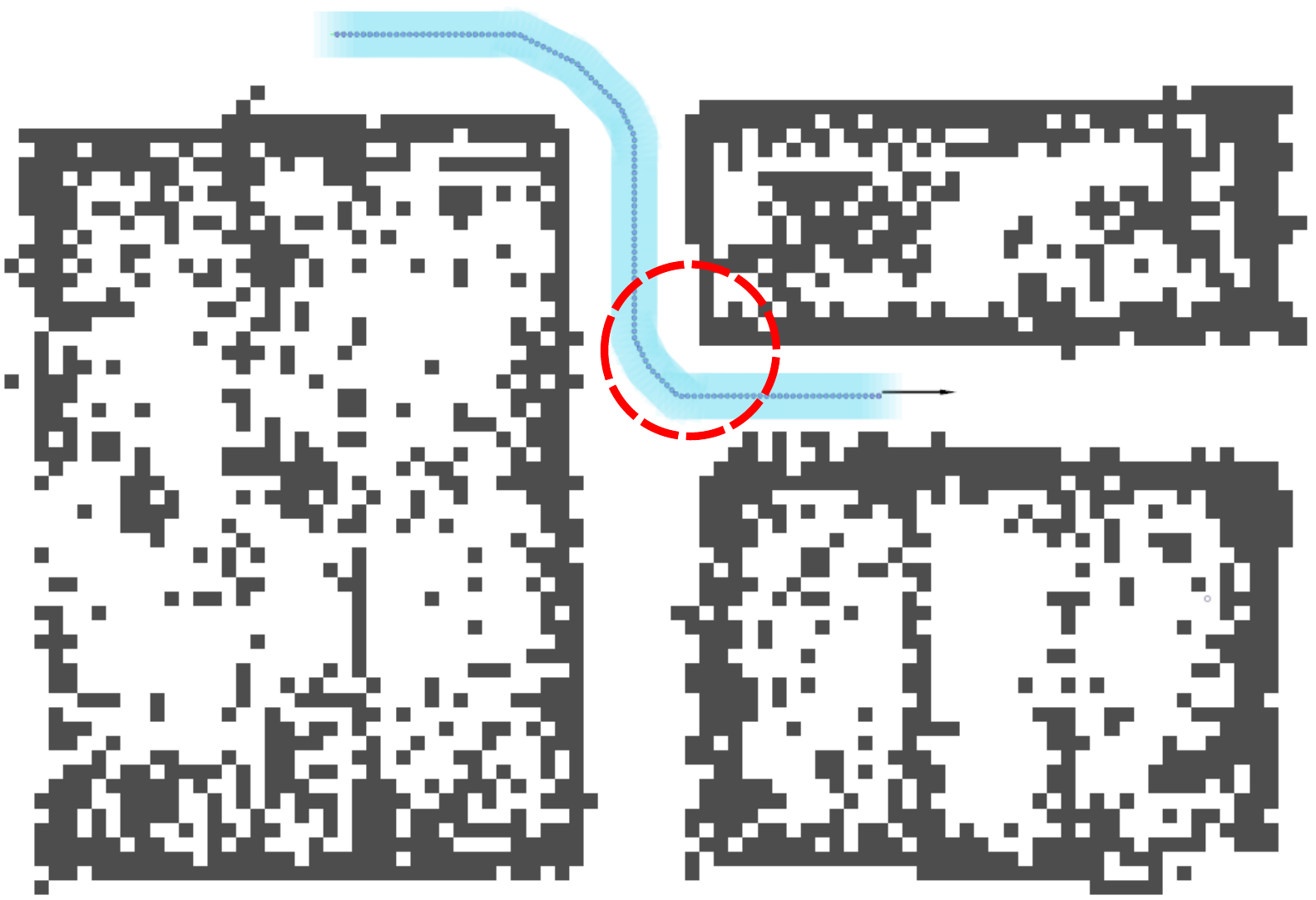}\label{fig:outdoor_lattice_path_3}}
	\centering
	\subfigure[]{\includegraphics[width=4cm]{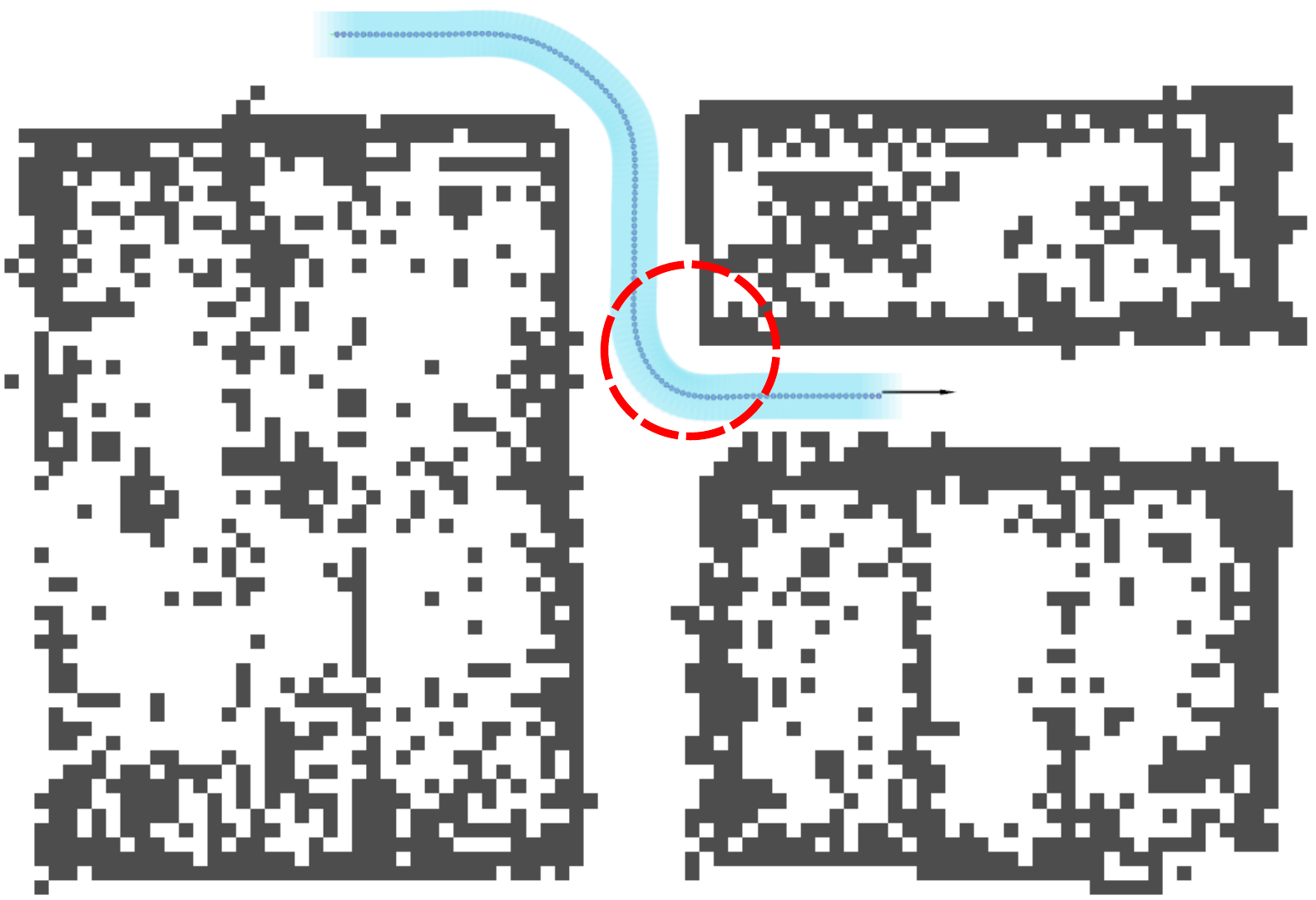}\label{fig:outdoor_lattice_path_4}}
	\caption{(a) The S-shaped environmental scenario. (b) The path obtained by the original state lattice-based path planner. (c) Smooth path with the path shown in (b) as input. (d) The path obtained by the proposed path planner. (e) Smooth path with the path shown in (d) as input. The light blue strips indicate the footprints of the robot along the path.}
	\label{fig:outdoor_lattice_path}
\end{figure}

\subsection{Experimental Setup}
As shown in Fig. \ref{fig:platform}, the mobile robot SUMMIT-XL HL is used as the experimental platform, which is equipped with a Velodyne VLP-16 LiDAR, an Intel Realsense D435i depth camera, and an Intel NUC computer. The footprint of the robot is approximated as a $ 0.8 $ $ \mathrm{m} \times 0.8 $ $ \mathrm{m} $ square, and the maximum linear velocity is $ 3.0 $ $ \mathrm{m/s} $. Considering the safety of robot navigation, the upper bound of the linear velocity is set to $ 1.5 $ $ \mathrm{m/s} $ in the experiments.

To generate a prior global map for autonomous navigation, we first employ a LiDAR-inertial odometry algorithm described in \cite{liosam2020shan} to build a large-scale 3-D point cloud map. And then, a point clouds segmentation algorithm presented in \cite{liu2021point} is used to filter out point clouds that hit the ground and tree canopy. Finally, the filtered point clouds are projected to a 2-D plane to derive traversable regions in the form of the occupancy grid map, as shown in Fig. \ref{fig:outdoor}. The dimension of the environment is approximately $ 210 $ $ \mathrm{m} \times 220 $ $ \mathrm{m} $, and the resolution of the grid map is $ 0.25 $ $\mathrm{m/cell}$. During robot navigation, the LiDAR-inertial odometry algorithm \cite{liosam2020shan} and the point clouds segmentation algorithm \cite{liu2021point} are also utilized to provide state estimation and local traversable regions for the robot, respectively.

\subsection{Comparison on Path Searching}
To validate that the proposed path searching approach can provide sufficient optimization margin for hard-constrained path smoothing approaches, we select an S-shaped scenario shown in Fig. \ref{fig:outdoor_env} to compare path searching approaches. The path shown in Fig. \ref{fig:outdoor_lattice_path_1} is obtained by the original state lattice-based path planner \cite{likhachev2009planning}. Since the path clearance is not considered in the original state lattice-based path planner, the searched path is close to the corner of the S-shaped turn, and the corresponding path vertices are fixed in the path smoothing process. As a result, the smoothed path is intuitively rough, as shown in Fig. \ref{fig:outdoor_lattice_path_2}. On the contrary, the path clearance is explicitly considered in the proposed path searching approach, thus the searched path shown in Fig. \ref{fig:outdoor_lattice_path_3} has a certain distance from obstacles and provides sufficient optimization margin for the path vertices near the corner. The final smoothed path shown in Fig. \ref{fig:outdoor_lattice_path_4} is much smoother than the path shown in Fig. \ref{fig:outdoor_lattice_path_2}, which demonstrates the proposed path searching approach can provide sufficient optimization margin for hard-constrained path smoothing approaches.

\begin{figure}[t]
	\centering
	\subfigure[]{\includegraphics[width=4cm]{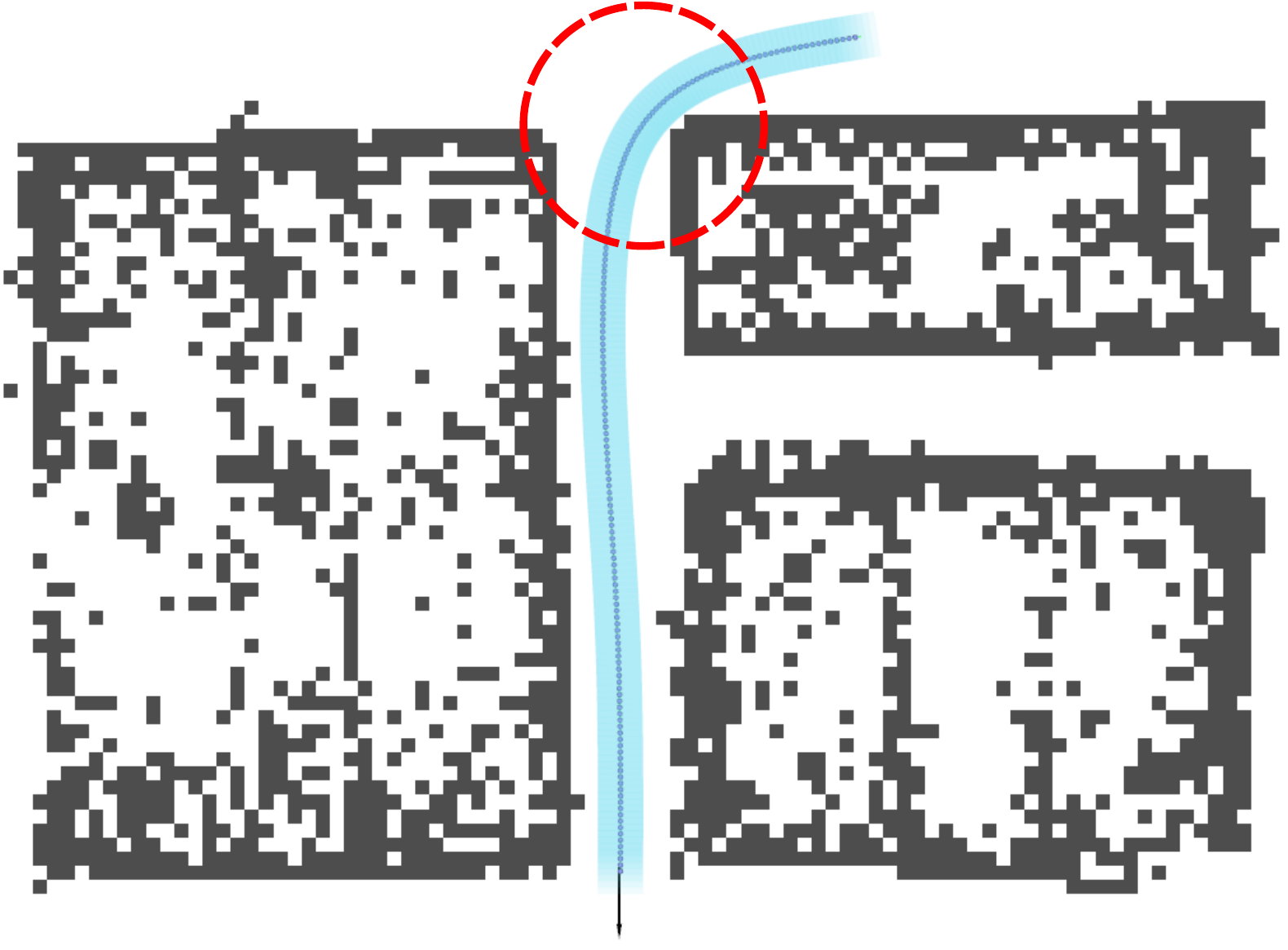}\label{fig:outdoor_lattice_path_5}}
	\centering
	\subfigure[]{\includegraphics[width=4cm]{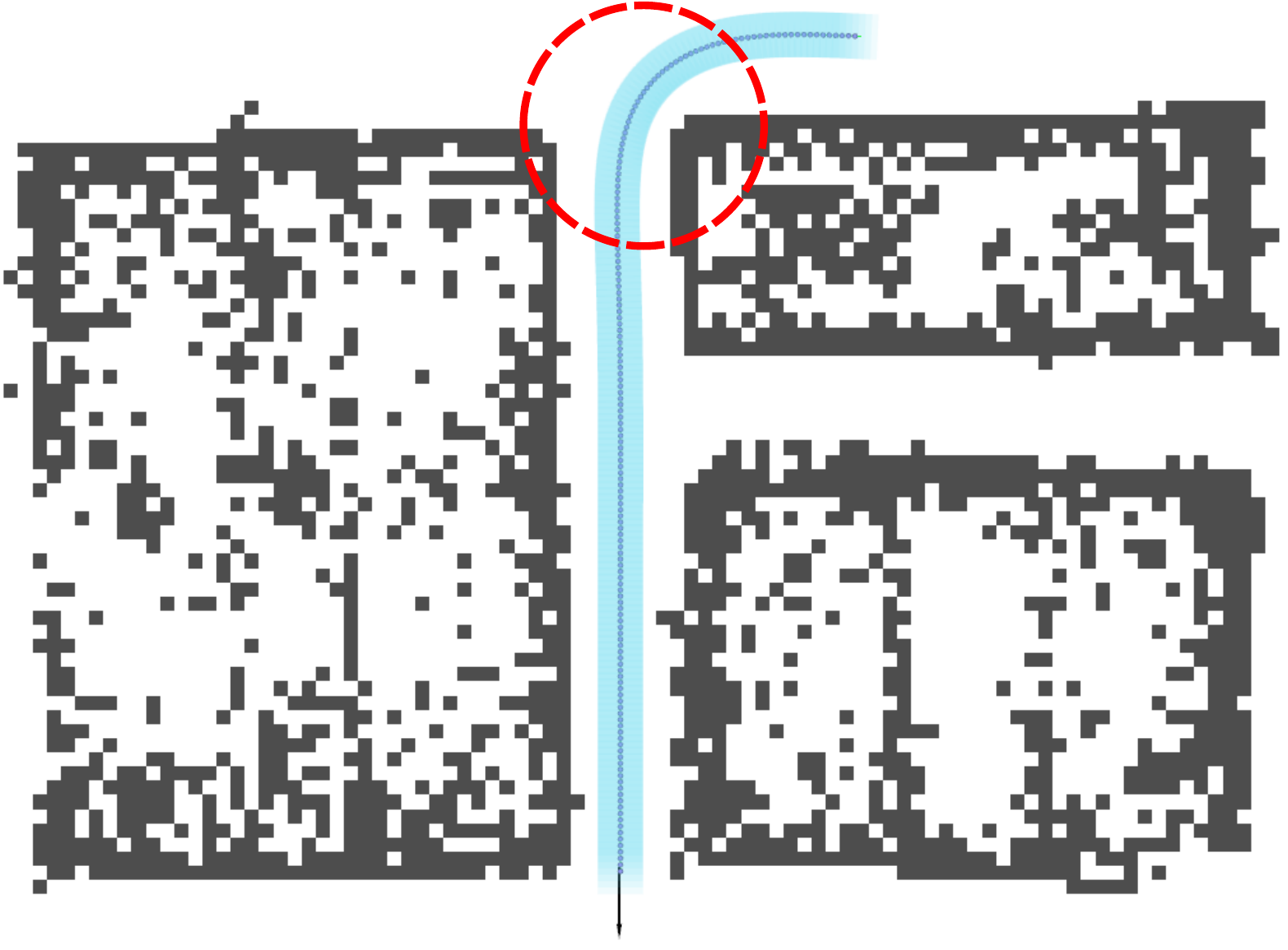}\label{fig:outdoor_lattice_path_6}}
	\caption{(a) Smooth path without penalizing the deviation from the reference path. (b) Smooth path with penalizing the deviation from the reference path. The light blue strips indicate the footprints of the robot along the path.}
	\label{fig:outdoor_smooth_path}
\end{figure}

\begin{table}[t]
	\centering
	\caption{Computational time (in millisecond) of path smoothing on the campus}
	\label{tbl:pathsmoothing}
	\begin{tabular}{ccccccc}
		\toprule
		\multicolumn{2}{c}{} 	   											& Mean     			& Max     			& Min     				& Std    			\\
		\midrule
		\multirow{3}{*}{Fig. \ref{fig:outdoor_lattice_path}} 	& TEB  		& 28.61				& 30.27        		&  27.10                & 0.72   			\\
																& SBA  		& 1.78				& 1.84        		&  1.74                 & 0.02   			\\
																& Ours 		& \textbf{0.37}		& \textbf{0.39}  	&  \textbf{0.37}  	   	& \textbf{0.01}   	\\
		\midrule
		\multirow{3}{*}{Fig. \ref{fig:outdoor_smooth_path}}   	& TEB 		& 31.24				& 32.73        		&  29.20        	   	& 0.84   			\\
																& SBA 		& 1.66				& 1.76        		&  1.62        	        & 0.03   			\\
																& Ours 		& \textbf{0.39}		& \textbf{0.40}  	&  \textbf{0.38}  	   	& \textbf{0.01}   	\\
		\bottomrule
	\end{tabular}
	\label{table:pathsmoothingtime_campus}
\end{table}

\subsection{Comparison on Path Smoothing}

\subsubsection{Comparison on Safety}
We select the start and goal configurations shown in Fig. \ref{fig:outdoor_smooth_path} to validate the performance improvement of the proposed path smoothing approach in terms of safety. For a fair comparison, the proposed path searching approach is used to generate the reference path for path smoothing. For testing purposes, we also do not limit the length of the initial path for path smoothing. Generally, hard-constrained path smoothing approaches treat all free space equally, namely, distance from feasible paths to obstacles is ignored. As a result, the optimized path is close to obstacles, as shown in Fig. \ref{fig:outdoor_lattice_path_5}. And the penalty of the deviation from the reference path is introduced in the proposed path smoothing approach. Because the reference path searched by the proposed path searching approach has a certain distance from obstacles, the safety of the path is implicitly considered during path smoothing, and the path clearance of the final optimized path is improved, as shown in Fig. \ref{fig:outdoor_lattice_path_6}. The minimum distance between the path shown in Fig. \ref{fig:outdoor_lattice_path_5} and obstacles is $ 0.56 $ $ \mathrm{m} $, which is the same as the circumscribed radius of the robot. While the path clearance of the path shown in Fig. \ref{fig:outdoor_lattice_path_6} is $ 0.75 $ $ \mathrm{m} $, and the path safety is improved by $ 33.9\% $.

\subsubsection{Comparison on Computational Efficiency}
We select the start and goal configurations shown in Figs. \ref{fig:outdoor_lattice_path} and \ref{fig:outdoor_smooth_path} to further compare the computational efficiency of path smoothing approaches. For a fair comparison, the proposed path searching approach is employed to generate the same reference path for SBA \cite{wen2020effmop}, TEB \cite{rosmann2017integrated}, and the proposed QP-based path smoothing approach. In each scenario, we also repeat the test $ 20 $ times. The statistics of the runtime performance are shown in Table \ref{table:pathsmoothingtime_campus}. Thanks to the convexity of the proposed path optimization formulation, the maximum runtime of the proposed path smoothing approach is less than $ 0.4 $ $ \mathrm{ms} $ in all two tests, and the computational efficiency of the proposed path smoothing approach is approximatively $ 4.4 $ and $ 73.6 $ times faster than that of SBA and TEB on average, respectively.

\begin{figure}[t]
	\centering
	\subfigure[Dealing with unknown static obstacles.]{
		\includegraphics[height=3cm]{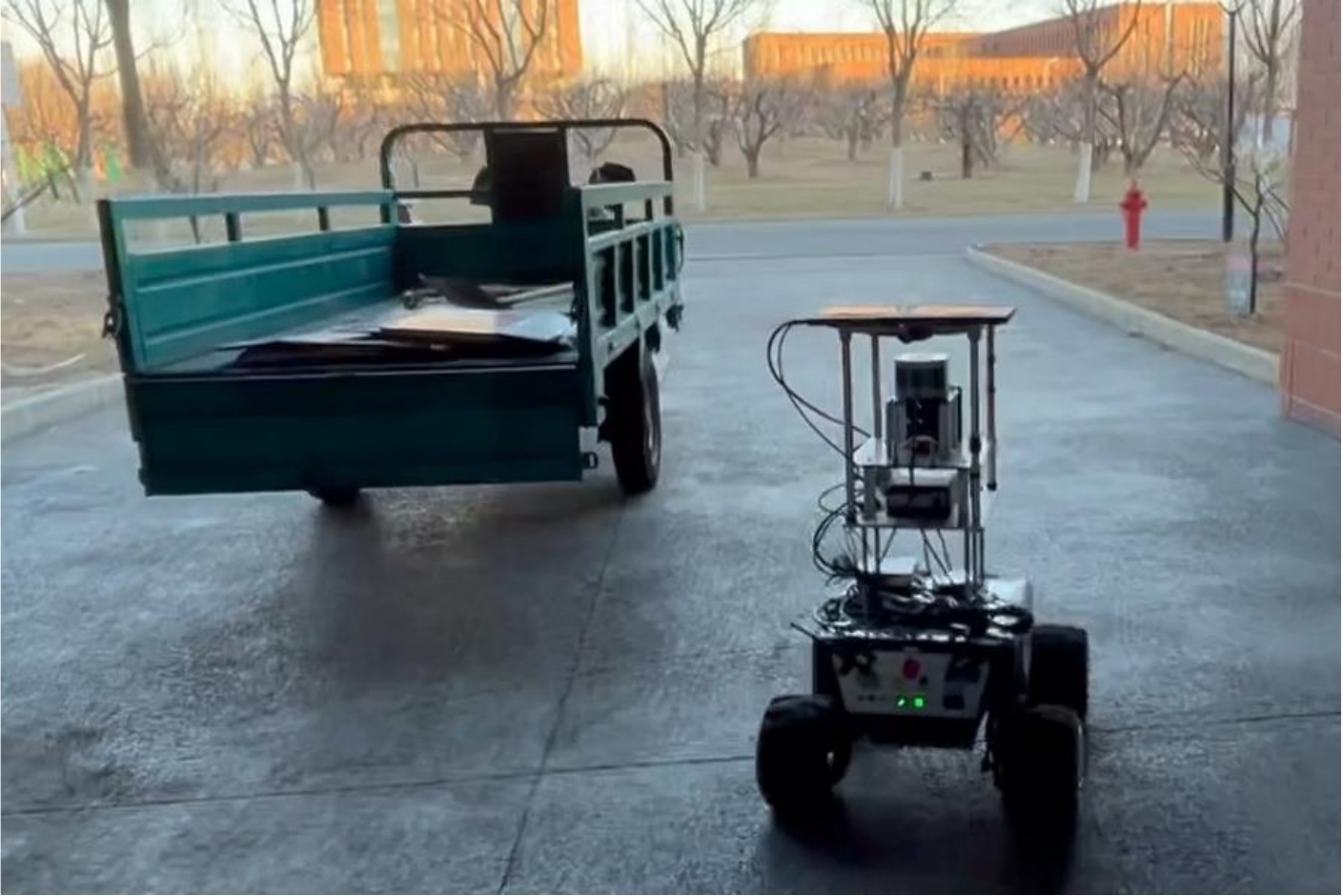}
		\includegraphics[height=3cm]{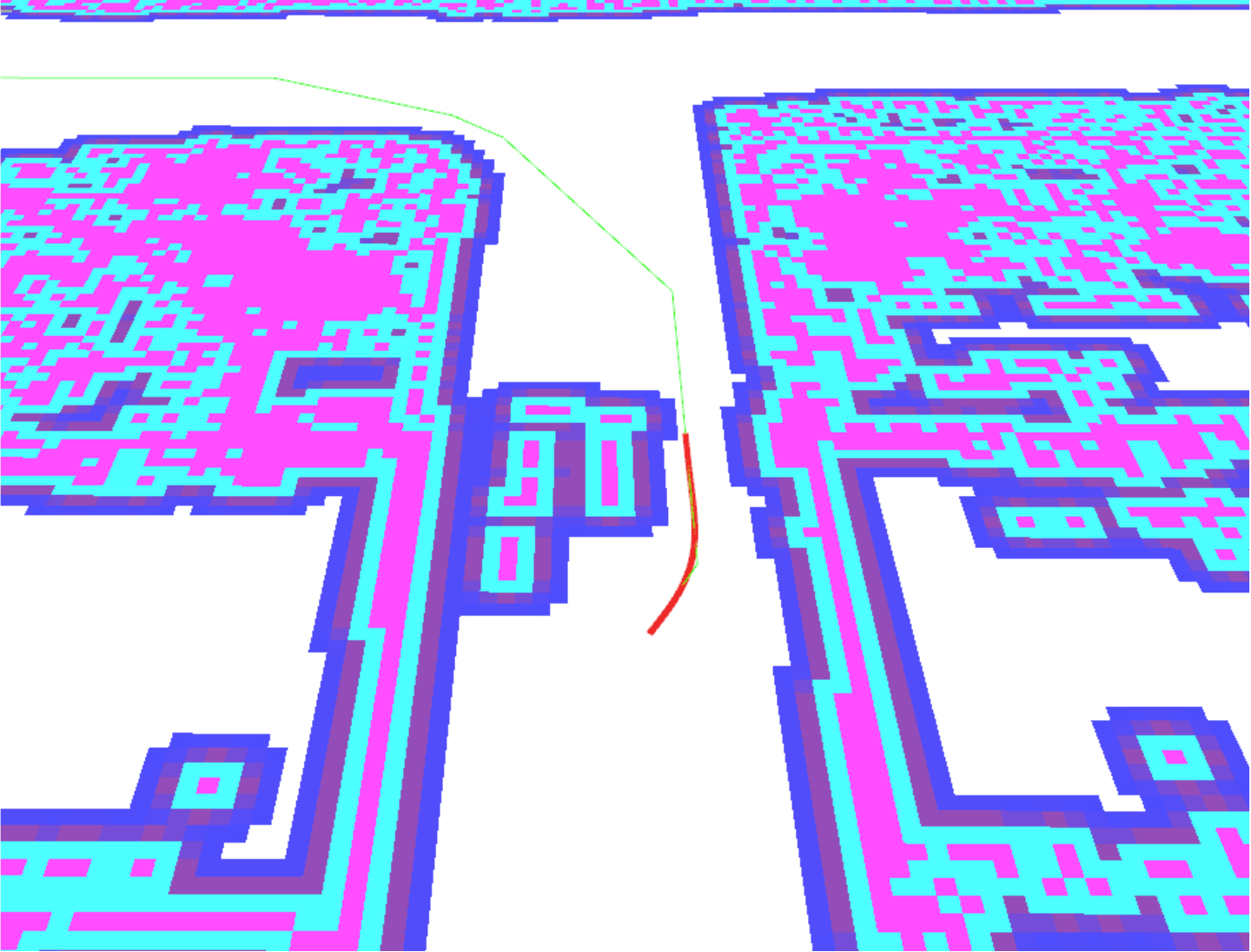}
		\label{fig:outdoor_navigation_1}
	}
	\centering
	\subfigure[Dealing with dynamic obstacles.]{
		\includegraphics[height=3cm]{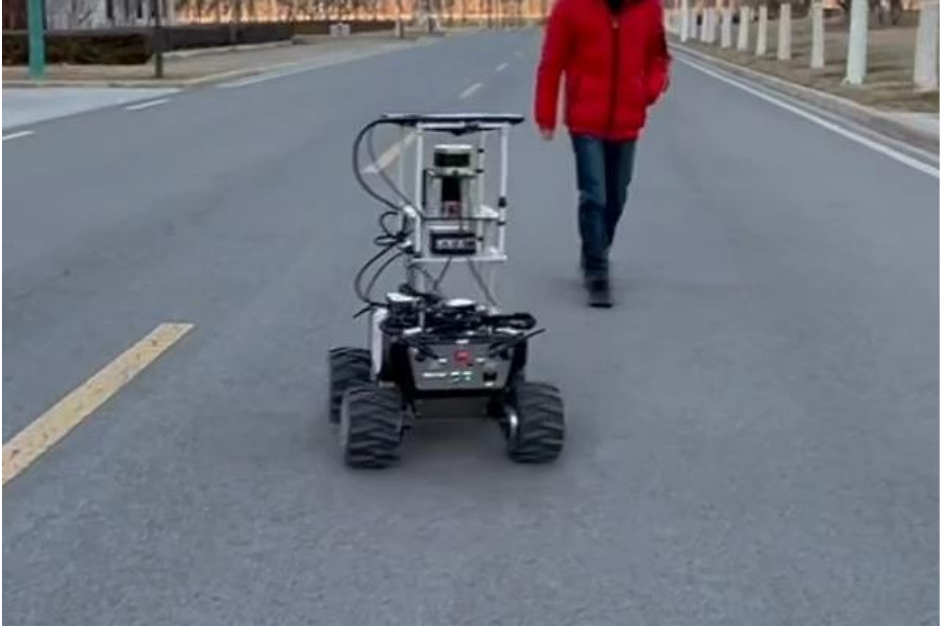}
		\includegraphics[height=3cm]{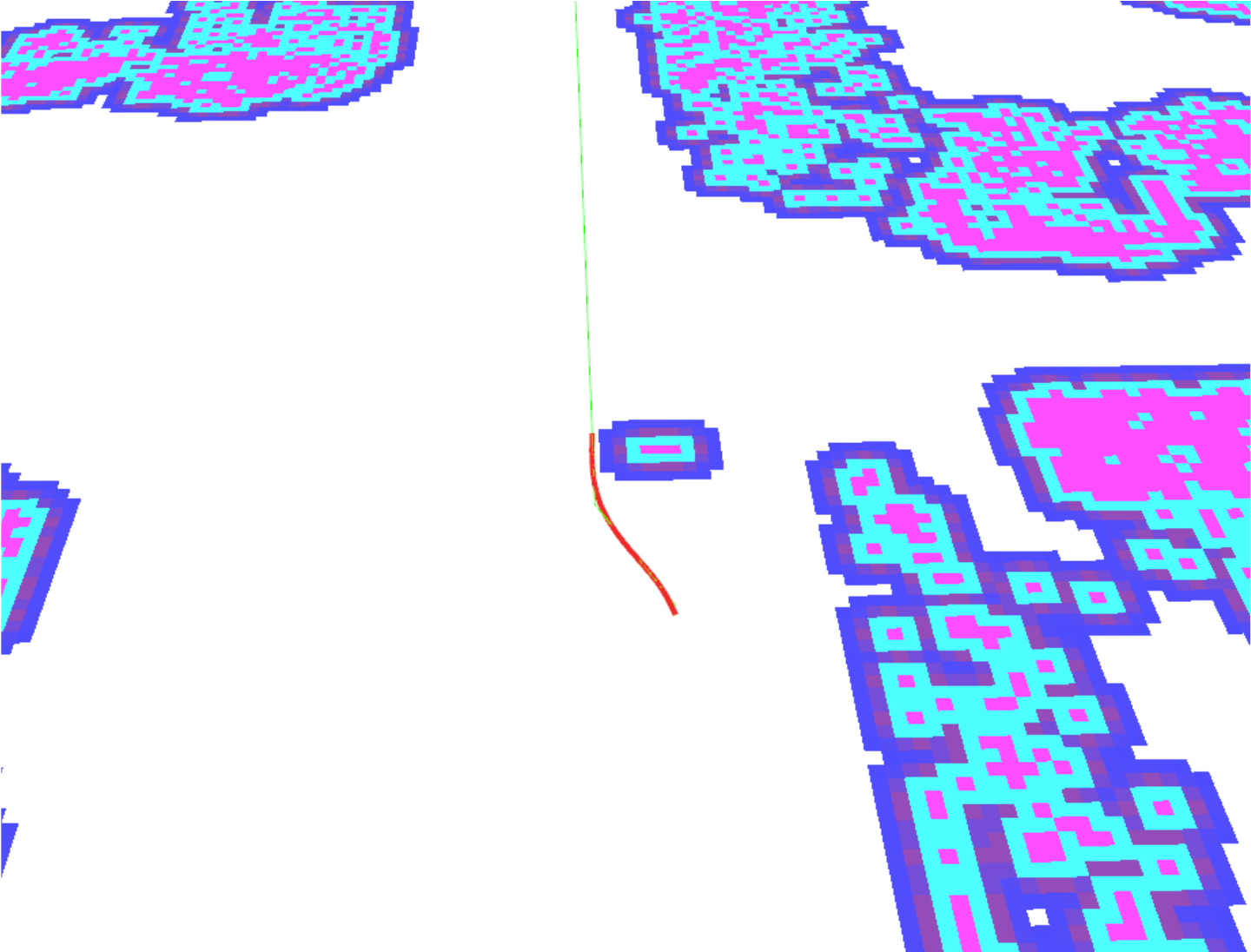}
		\label{fig:outdoor_navigation_3}
	}
	\caption{Screenshots of outdoor navigation results. The green curve denotes the path generated by the proposed path searching approach, and the red curve represents the path smoothed by the proposed path smoothing approach. The pink cells denote obstacles, and the blue cells represent the inflation cells of obstacles according to the footprint of the robot.}
\end{figure}

\subsection{Autonomous Navigation}
Finally, the proposed path searching and path smoothing approaches are integrated into the powerful ROS navigation stack\footnote{\url{https://github.com/ros-planning/navigation}} to validate the effectiveness and practicability of the G\ensuremath{^2}VD planner. In particular, we implement a G\ensuremath{^2}VD cost map layer on top of the original ROS cost map. On this basis, the proposed path searching approach and the path smoothing approach are implemented as global and local planner plugins for the ROS navigation stack, respectively. Considering the computational efficiency and sensing range, the length of the initial local path is set to $4.0$ $\mathrm{m}$. As illustrated in Fig. \ref{fig:outdoor}, we select two sets of different start and goal configurations for outdoor navigation. The total travel distances of these two sets of outdoor navigation are approximately $ 118.3 $ $ \mathrm{m} $ and $ 165.9 $ $ \mathrm{m} $, respectively.

Here we summarize several representative experimental results of outdoor autonomous navigation to demonstrate the key characteristics of the G\ensuremath{^2}VD planner. More details are included in the video.

\subsubsection{Dealing With Static Obstacles}
Fig. \ref{fig:outdoor_navigation_1} illustrates the scenario with unknown static obstacles. The robot avoids a temporarily parked tricycle smoothly, according to the reliable path smoothing results.

\subsubsection{Dealing With Dynamic Obstacles}
Fig. \ref{fig:outdoor_navigation_3} shows the scenario with dynamic obstacles. The robot implements fast re-planning and avoids an oncoming person successfully, thanks to the efficient path smoothing approach.

\textit{Remark:} In this work, dynamic obstacles are regarded as instantaneous static obstacles. Namely, both dynamic and static obstacles are considered in a unified framework. Moving dynamic obstacles and unknown static obstacles are detected by external sensors mounted on the robot and the states of affected cells in the G\ensuremath{^2}VD are updated according to the update mechanism described in Section \ref{update_mechanism}.

\section{Conclusion}
\label{conclusion}
In this paper, an efficient motion planning approach called G\ensuremath{^2}VD planner is newly proposed for mobile robots. Based on a G\ensuremath{^2}VD, a new state lattice-based path planner is proposed, in which the search space is reduced to a Voronoi corridor to further improve the search efficiency. And an efficient QP-based path smoothing approach is presented, wherein the clearance to obstacles is considered to improve the path clearance of hard-constrained path smoothing approaches. We validate the G\ensuremath{^2}VD planner in various complex simulation scenarios and outdoor environments. The results show that the computational efficiency is improved by 17.1\% in the path searching stage, and path smoothing with the proposed approach is 25.3 times faster than an advanced sparse-banded structure-based path smoothing approach.

In the future, we plan to integrate the intention/trajectory prediction of pedestrians/vehicles into the proposed framework to enhance the foreseeability of the motion planner.

\bibliographystyle{IEEEtran}
\bibliography{ref}

\begin{IEEEbiography}[{\includegraphics[width=1in,height=1.25in,clip,keepaspectratio]{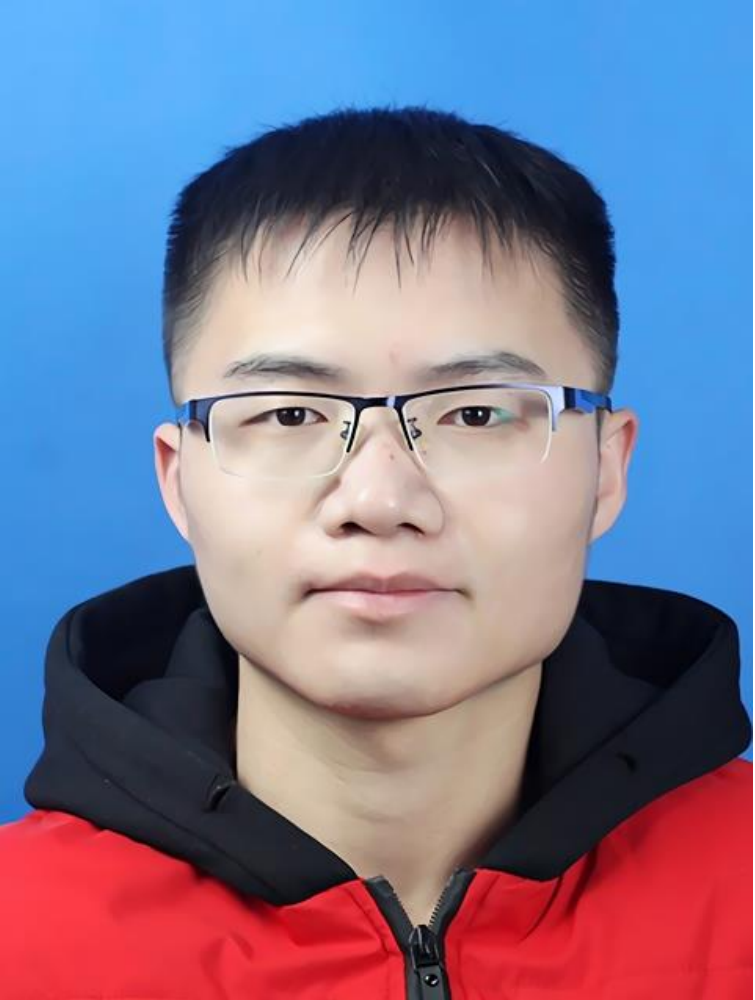}}]{Jian Wen} received the B.S. degree in automation and the Ph.D. degree in control science and engineering from Nankai University, Tianjin, China, in 2017 and 2022, respectively. 
	
He is currently working as a Senior Algorithm Engineer with the Group of Autonomous Driving, Xiaomi EV Company Limited, Beijing, China. His research interests include behavioral reasoning, decision making, and motion planning for autonomous driving vehicles.
\end{IEEEbiography}

\begin{IEEEbiography}[{\includegraphics[width=1in,height=1.25in,clip,keepaspectratio]{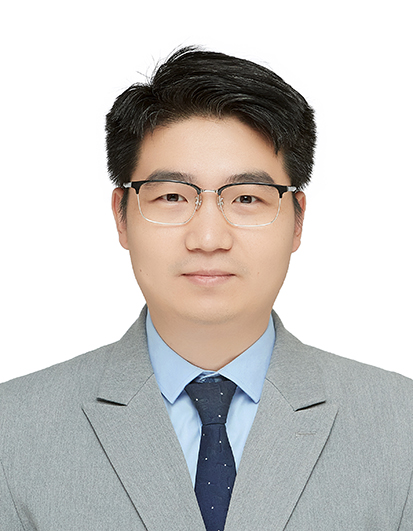}}]{Xuebo Zhang} (M'12$-$SM'17) received the B.Eng. degree in automation from Tianjin University, Tianjin, China, in 2006, and the Ph.D. degree in control theory and control engineering from Nankai University, Tianjin, China, in 2011.
	
From 2014 to 2015, he was a Visiting Scholar with the Department of Electrical and Computer Engineering, University of Windsor, Windsor, ON, Canada. He was a Visiting Scholar with the Department of Mechanical and Biomedical Engineering, City University of Hong Kong, Hong Kong, in 2017. He is currently a Professor with the Institute of Robotics and Automatic Information System, Nankai University, and Tianjin Key Laboratory of Intelligent Robotics, Nankai University. His research interests include planning and control of autonomous robotics and mechatronic system with focus on time-optimal planning and visual servo control; intelligent perception including robot vision, visual sensor networks, SLAM; reinforcement learning and game theory.
	
Dr. Zhang is a Technical Editor of \emph{IEEE/ASME Transactions on Mechatronics} and an Associate Editor of \emph{ASME Journal of Dynamic Systems, Measurement, and Control}.
\end{IEEEbiography}

\begin{IEEEbiography}[{\includegraphics[width=1in,height=1.25in,clip,keepaspectratio]{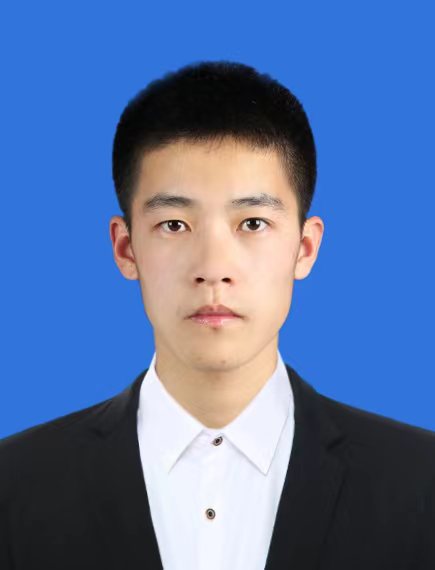}}]{Qingchen Bi} received the B.S. degree in detection guidance and control technology from Northwestern Polytechnical University, Xi’an, China, in 2021, where he is currently pursuing the Ph.D. degree in control science and engineering with the Institute of Robotics and Automatic Information System, Nankai University, Tianjin, China.
	
His research interests include mobile robot motion planning and autonomous exploration.
\end{IEEEbiography}

\begin{IEEEbiography}[{\includegraphics[width=1in,height=1.25in,clip,keepaspectratio]{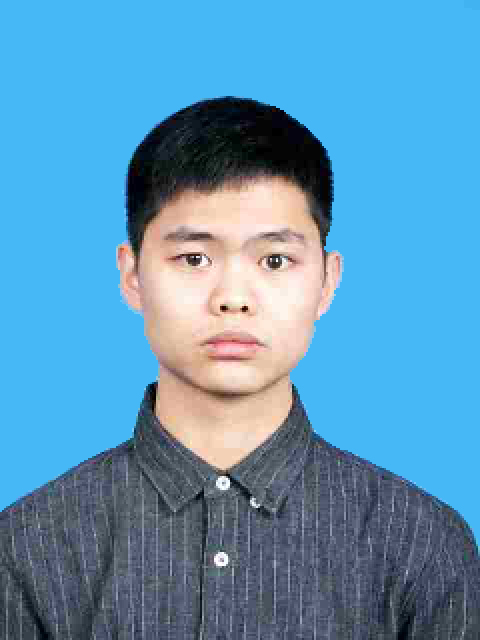}}]{Hui Liu} received the B.S. degree in intelligence science and technology from Hebei University of Technology, Tianjin, China, in 2019, and the M.S. degree in control science and engineering from Nankai University, Tianjin, China, in 2022. 
	
He is currently working as an Algorithm Engineer at Baidu Inc., Beijing, China. His research interests include SLAM and calibration.
\end{IEEEbiography}

\begin{IEEEbiography}[{\includegraphics[width=1in,height=1.25in,clip,keepaspectratio]{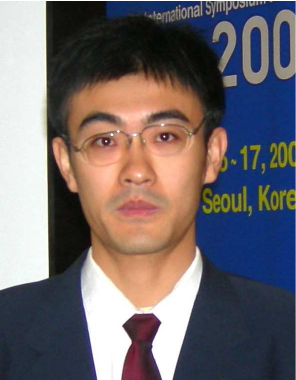}}]{Jing Yuan} (M'12) received the B.S. degree in automatic control, and the Ph.D. degree in control theory and control engineering from Nankai University, Tianjin, China, in 2002 and 2007, respectively.
	
Since 2007, he has been with the College of Computer and Control Engineering, Nankai University, where he is currently a Professor. His current research interests include robotic control, motion planning, and SLAM.
\end{IEEEbiography}

\begin{IEEEbiography}[{\includegraphics[width=1in,height=1.25in,clip,keepaspectratio]{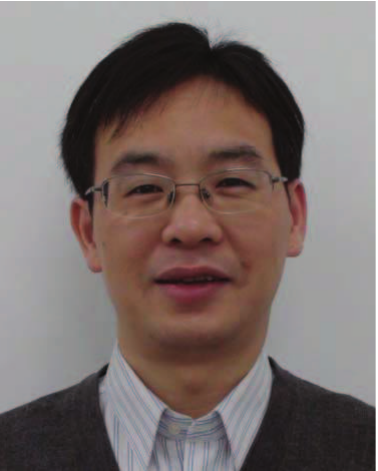}}]{Yongchun Fang} (S'00$-$M'02$-$SM'08) received the B.S. degree in electrical engineering and the M.S. degree in control theory and applications from Zhejiang University, Hangzhou, China, in 1996 and 1999, respectively, and the Ph.D. degree in electrical engineering from Clemson University, Clemson, SC, in 2002.
	
From 2002 to 2003, he was a Post-Doctoral Fellow with the Sibley School of Mechanical and Aerospace Engineering, Cornell University, Ithaca, NY, USA. He is currently a Professor with the Institute of Robotics and Automatic Information System, Nankai University, Tianjin, China. His research interests include visual servoing, AFM-based nano-systems, and control of underactuated systems including overhead cranes.
\end{IEEEbiography}

\end{document}